\documentclass{article}

\usepackage[final,nonatbib]{neurips_2023}

\usepackage[utf8]{inputenc} %
\usepackage[T1]{fontenc}    %
\usepackage{url}            %
\usepackage{booktabs}       %
\usepackage{amsfonts}       %
\usepackage{nicefrac}       %
\usepackage{microtype}      %

\usepackage{graphicx}
\usepackage{amsmath}
\usepackage[dvipsnames,table,xcdraw]{xcolor}
\definecolor{citecolor}{RGB}{34,139,34}
\usepackage[pagebackref,breaklinks,colorlinks,citecolor=citecolor,bookmarks=false,hypertexnames=false]{hyperref}
\usepackage{enumitem}
\usepackage[font=small,labelfont=bf]{caption}
\usepackage{subcaption}

\graphicspath{{figures/}}
\graphicspath{{figures_supp/}}

\newcommand{\B}[1]{{\textbf{#1}}}
\newcommand{\SC}[1]{{\textsc{#1}}}
\newcommand{\commentout}[1]{}
\newcommand{\aprox}{\raisebox{0.5ex}{\texttildelow}}

\newcommand{\refeqn}[1]{Eqn.~\ref{#1}}
\newcommand{\reffig}[1]{Fig.~\ref{#1}}
\newcommand{\reftbl}[1]{Table~\ref{#1}}
\newcommand{\refsec}[1]{Sec.~\ref{#1}}

\def\Acronym{EgoEnv}

\title{EgoEnv: Human-centric environment representations from egocentric video}

\author{
        Tushar Nagarajan$^{2}$, 
        Santhosh Kumar Ramakrishnan$^1$,
        Ruta Desai$^2$, \\
        \textbf{James Hillis$^2$},
        \textbf{Kristen Grauman$^{1,2}$} \\
        $^1$University of Texas at Austin, $^2$FAIR, Meta
}

\begin{document}

\maketitle

\begin{abstract}
First-person video highlights a camera-wearer's activities \emph{in the context of their persistent environment}. However, current video understanding approaches reason over visual features from short video clips that are detached from the underlying physical space and  capture only what is immediately visible. To facilitate human-centric environment understanding, we present an approach that links egocentric video and the environment by learning representations that are predictive of the camera-wearer's (potentially unseen) local surroundings. We train such models using videos from agents in simulated 3D environments where the environment is fully observable, and test them on human-captured real-world videos from unseen environments. On two human-centric video tasks, we show that models equipped with our environment-aware features consistently outperform their counterparts with traditional clip features. Moreover, despite being trained exclusively on simulated videos, our approach successfully handles real-world videos from HouseTours and Ego4D, and achieves state-of-the-art results on the Ego4D NLQ challenge. Project page: \url{https://vision.cs.utexas.edu/projects/ego-env/}
\end{abstract}

\section{Introduction}\label{sec:introduction}
Egocentric video offers a unique view into human activities through the eyes of a camera-wearer. Understanding this type of video is core to building augmented reality (AR) applications that can provide context-relevant assistance to humans based on their activity.
Ego-video is thus the subject of several recent datasets and benchmarks that are driving new research~\cite{li2015delving,sigurdsson2018charades,damen2022rescaling,grauman2022ego4d}.  

\begin{figure}[t]
\centering
\includegraphics[width=\linewidth]{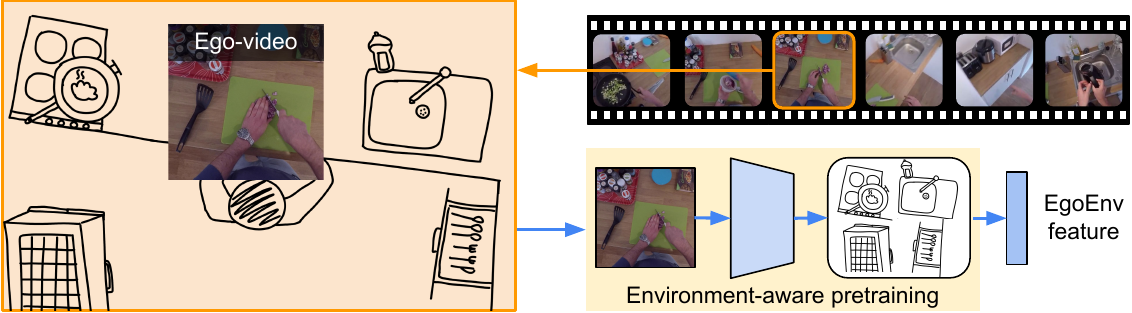}
\vspace*{-0.2in}
\caption{\textbf{Main idea.} Video models trained on short egocentric clips capture a narrow, instantaneous view of human activity (e.g., cutting an onion at the counter) without considering the broader context to which the activity is tied (e.g., the pan to the left to cook the onions, the fridge further behind to store leftovers). We propose to ground video in its underlying 3D environment by learning representations that are predictive of their surroundings, thereby enhancing standard clip-based models with complementary \emph{environment} information. 
}
\vspace*{-0.2in}
\label{fig:concept}
\end{figure}

A key feature of the egocentric setting is the tight coupling of a camera-wearer and their persistent physical environment, i.e., a person's mental model of their surroundings informs their actions. This mental model is important, for example, to reach for a cabinet door out of view, to re-visit the couch to search for a misplaced phone or to visit spaces configured to support certain activities. This raises an important need for human-centric environment understanding --- to learn representations from video that capture the camera-wearer's activities \emph{in the context of their environment}. Such representations would encode the human-environment link, and allow models to jointly reason about both (e.g., to answer ``what did the person cut near the sink?''). See \reffig{fig:concept}.

Despite its importance, there has been only limited work on learning human-centric environment representations. Current video models segment a video into short clips (1-2s long) and then aggregate clip features over time (e.g., with recurrent, graph or transformer-based networks) for tasks like action forecasting~\cite{furnari2019would,grauman2022ego4d,nagarajan2020egotopo,girdhar2021anticipative}, temporal action localization~\cite{zhang2020learning,lin2019bmn,liu2021context,zhao2021video}, episodic memory~\cite{grauman2022ego4d,datta2022emqa}, and movie understanding~\cite{wu2021towards}.
Critically, the clip features encode what is immediately visible in a short time window, and their aggregation over \emph{time} does not equate to linking them in \emph{physical space}. Other approaches use explicit camera pose information (e.g., from SLAM) to localize the camera-wearer, but not its relation to surrounding objects (e.g., forecasting~\cite{park2016egocentric,rhinehart2017first,guan2020generative}) or to group activities by location, but do not learn representations for agent video (e.g., affordance prediction~\cite{rhinehart2016learning,nagarajan2020egotopo}).

To address these shortcomings, we propose to learn environment-aware video representations that encode the surrounding physical space. Specifically, we define the \emph{local environment state} at each time-step of an egocentric video as the set of objects (and their rough distance) in front, to the left, right, and behind the camera-wearer. See \reffig{fig:local_state}.
We use this state as supervision to train a transformer-based video encoder model that aggregates visual information across a video to build an \emph{environment memory}, which can be queried to predict the local state at any point in the video. 

The local state captures the rough layout of objects relative to the camera-wearer. It is important for understanding physical space --- it provides a semantic signal to localize the camera-wearer (e.g., in the living room, from the arrangement of couches, lamps and tables) --- as well as human behavior, since people move towards  layouts that support activities (e.g., stove-top areas, dressers). Predicting the local state thus involves capturing the natural statistics of object layouts across different homes and then translating contexts across environments to reason about new ones. Once trained, given an observation from a new video, our model produces a drop-in \Acronym~feature which encodes \emph{environment} information to complement the \emph{action} information in existing video clip features.

An important practical question is how to supervise such a representation.
Sourcing local state labels requires agent and object positions and omnidirectional visibility at each time step. This is challenging as egocentric videos only offer sparse coverage of the environment. Furthermore, they are prone to object detection, tracking, and SLAM failures.
Therefore, for training we turn to videos generated by agents in simulation. %
This allows us to sample diverse, large-scale trajectories to cover the environment, while also providing ground-truth local state. 
Once trained with simulated video, we apply our models to \emph{real-world} videos from new, unseen environments. 

We demonstrate our EgoEnv approach on two video tasks where joint reasoning of both human action and the underlying physical space is required: (1) inferring the room category that the camera wearer is physically in as they move through their environment, and (2) localizing the answer to a natural language query in an egocentric video.
These tasks support many potential applications, including AR systems that can offer context-relevant assistance.

We are the first to demonstrate the value of 3D simulation data for real-world ego-video understanding. 
Our experiments show that by capitalizing on both geometric and semantic cues in our proposed ``local environment state'' task, we can leverage video walkthroughs from \emph{simulated agents} in HM3D scenes~\cite{ramakrishnan2021hm3d} to ultimately enable downstream human-centric environment models on \emph{real-world} videos from HouseTours~\cite{chang2020semantic} and Ego4D~\cite{grauman2022ego4d}. Furthermore, models equipped with our \Acronym~features outperform both popular scene classification~\cite{zhou2017places} and natural language video localization models~\cite{zhang2020learning}, and achieve state-of-the-art results on the Ego4D NLQ challenge leaderboard.
\section{Related work}

\vspace{-0.05in}
\paragraph{Video understanding in 3D environments.}
Prior work encodes short video clips~\cite{carreira2017quo,wang2016temporal,pirri2019anticipation,hara2018can} or temporally aggregates them %
for additional context~\cite{wu2019long,furnari2019would,zhang2020learning,lin2019bmn,liu2021context,zhao2021video,wu2021towards}. However, these methods treat the video as a temporal sequence and fail to capture the spatial context from the underlying persistent environment. 
For egocentric video, prior work has used structure from motion (SfM) to map people and objects for trajectory~\cite{park2016egocentric} and activity forecasting~\cite{guan2020generative} and action grounding in 3D~\cite{rhinehart2016learning,damen2016you}. 
These approaches localize the camera-wearer but do not learn representations for the camera-wearer's %
surroundings.
The model of \cite{liu2022egocentric} associates features to voxel maps to localize actions; however, they require a pre-computed 3D scan of the environment at both training and inference. Prior work groups clips by rough spatial location as topological graphs~\cite{nagarajan2020egotopo} or activity threads~\cite{price2022unweavenet}, but they stop short of learning representations using these groups. In contrast, we %
explicitly learn features that relate clips based on their spatial layout, for each step of an ego-video.

\vspace{-0.05in}
\paragraph{Video representation learning.}
Traditional video understanding methods learn representations by training models on large, manually curated video action recognition datasets~\cite{carreira2017quo,miech2019howto100m,grauman2022ego4d}. Recent self-supervised learning (SSL) approaches eliminate the supervision requirement by leveraging implicit temporal signals~\cite{wei2018learning,misra2016shuffle,van2018representation,jayaraman2015learning,han2019video,vondrick2016anticipating,wang2017transitive,wang2019learning,recasens2021broaden,feichtenhofer2021large}. 
In contrast, we learn features that encode the local spatial-state of the environment (as opposed to temporal signals).  Further, we show how to leverage state information that is readily accessible in simulation, but not in video datasets (i.e., locations and semantic classes of objects surrounding the agent) for training. 

\vspace{-0.05in}
\paragraph{Environment features for embodied AI.} 
In embodied AI, pose-estimates are used to build maps~\cite{chen2019learning,chaplot2020learning,ramakrishnan2020occupancy}, as edge features in graphs~\cite{chaplot2020neural,chang2020semantic}, as spatial embeddings for episodic memories~\cite{fang2019scene}, to project features to a grid map~\cite{gupta2017cognitive,henriques2018mapnet,cartillier2021semantic} or to learn environment features for visual navigation~\cite{ramakrishnan2021epc}. However, these approaches are explored solely in simulation and typically require accurate pose-estimates or smooth action spaces, and thus are not directly applicable to egocentric videos.
Research on world-models~\cite{dosovitskiy2017learning,ha2018world,hafner2019dream} and unseen panorama reconstruction~\cite{jayaraman2018learning,song2018im2pano3d,koh2021pathdreamer} \emph{hallucinate} the effect of agent actions to aid decision-making in simulation. %
In contrast, we aim to learn environment features for an egocentric camera-wearer to aid real-world video understanding. 

\vspace{-0.05in}
\paragraph{Learning from simulated data.} 
Prior work has proposed cost-effective ways to generate large-scale synthetic image datasets for various vision tasks~\cite{su2015render,tobin2017domain,denninger2019blenderproc,hinterstoisser2018pre,eftekhar2021omnidata,roberts2021hypersim}.
In robotics, simulation environments have been developed to quickly and safely train policies, with the eventual goal of transferring them to real world applications~\cite{koenig2004design,todorov2012mujoco,kolve2017ai2,chang2017matterport3d,xia2018gibson,savva2019habitat,ramakrishnan2021hm3d,szot2021habitat}. The resulting \emph{sim-to-real} problem, where models must adapt to changes in simulator and real-world domains, is an active area of research for robot navigation~\cite{tobin2017domain,kadian2020sim2real,rosano2021embodied,bigazzi2021out,anderson2021sim}. However, simulated data for video understanding is much less explored. Prior work has synthesized data for human body pose estimation~\cite{chen2016synthesizing,varol2017learning,yuan20183d,doersch2019sim2real}, trajectory forecasting~\cite{liang2020simaug}, and action recognition~\cite{roberto2017procedural}. 
Rather than model human behavior, our approach is the first to directly capture the environment surrounding the camera-wearer for real-world video understanding tasks.

\section{Approach} \label{sec:approach}

Our goal is to learn \Acronym~representations that encode the local surroundings of the camera-wearer. %
Such a representation would implicitly maintain a semantic memory of surrounding objects beyond what is immediately visible, and coupled with a standard video feature, would allow models to jointly reason about activities and the underlying physical space.  Directly appending %
the camera pose with each video frame may capture the local state; however noise in pose estimated from ego-video with quick head motions and characteristic blur limits its utility (see Supp. for experiments). 

Instead, we introduce an approach that leverages simulated environments where perfect state information is available to train models that can link visual information to the physical surroundings. To this end, we first define the local state prediction task in simulation (\refsec{sec:local_state}). Next, we introduce our transformer-based architecture that predicts the local state in videos (\refsec{sec:pretraining}). Finally, we show how our model trained in simulation generates \emph{environment features} for real-world egocentric video frames (\refsec{sec:downstream}).

\begin{figure*}[t]
\centering
\includegraphics[width=\linewidth]{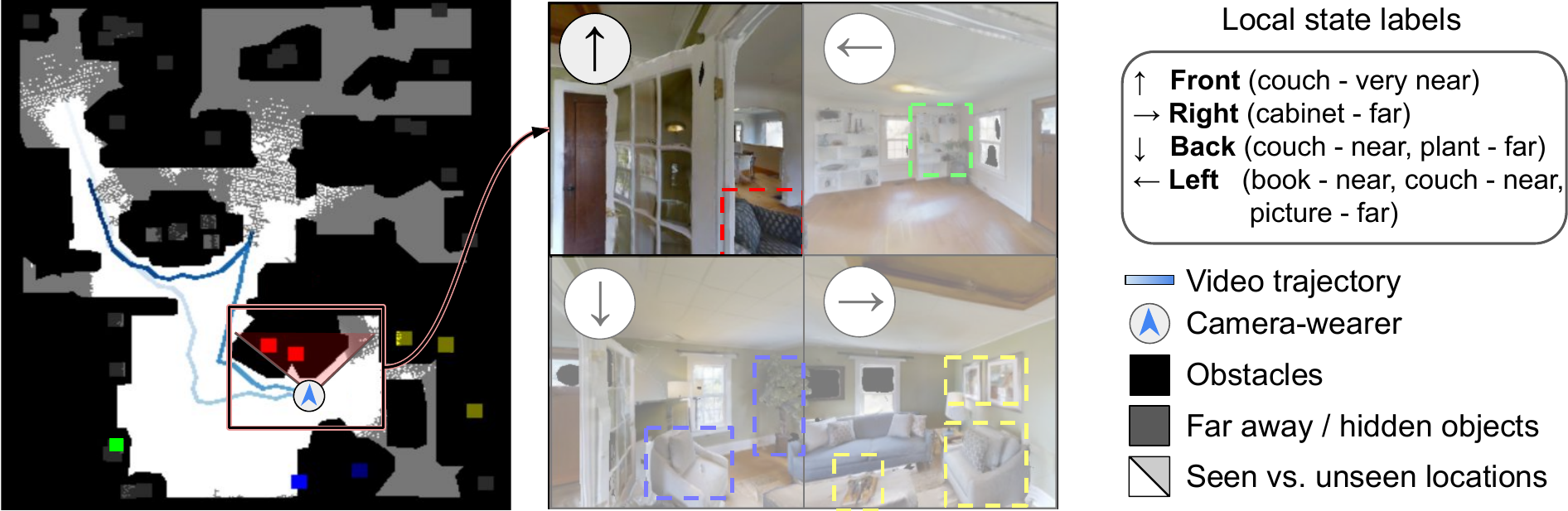}
\vspace*{-0.2in}
\caption{
\textbf{Local environment state task.} 
\textbf{Left panel:} Top-down environment view showing the camera-wearer path (blue gradient line) and nearby objects (colored boxes). 
\textbf{Middle panel:} Egocentric view in each direction. Only the forward view (top left) is observed. Remaining views are shown for clarity.
\textbf{Task:} Given an egocentric video of a person in their environment, the model must predict the set of objects (and their rough distance) in front, to the left, right, and behind the camera-wearer at each time-step (right panel). The model only sees the forward ego-view (middle panel, top left) and does not have access to the top-down map.
Note that not all parts of the environment are seen during a walkthrough (white vs. grey regions on map) --- models must link seen observations based on their shared space, as well as anticipate unseen surroundings based on statistics of training environments. Best viewed in color. 
}
\label{fig:local_state}
\end{figure*}

\vspace{-0.05in}
\subsection{Local environment state} \label{sec:local_state}

We require a model that is aware of not just what is immediately visible in a single frame, but also of the camera-wearer's surroundings. We therefore define the \emph{local environment state} of the camera-wearer as the set of objects in each relative direction --- i.e., what objects are to the front, left, right or behind the camera-wearer, together with their rough distance from the camera-wearer --- and train a model to predict this state. 
Our definition of local state takes inspiration from cognitive science~\cite{han2012objects,krokos2019virtual},
and offers supervision signals that are both geometric (relative object locations) and semantic (semantic object labels), which we observe leads to strong representations.

More formally, let $\mathcal{O}$ be a set of object classes. For a frame $f$ from a video trajectory 
in an environment, the local state is a tuple $(y_o, y_r)$. $y_o$ is a $4 \times |\mathcal{O}|$ dimensional matrix which represents instances of each object class in the four cardinal directions relative to the camera-wearer. $y_r$ is a matrix of the same size containing the distance of the objects in $y_o$ from the camera-wearer, discretized into $5$ distance ranges between $0.25-5.0m$\footnote{For object classes with multiple instances, we select the nearest one.}. For direction $i$ and object class $j$, the labels are:
\begin{align}
y_o[i,j] &= 
\begin{cases}
    1 & \text{if~} d(p_a, p_j) < \delta \land \theta(p_a, p_j) = i \\
    0 & \text{otherwise},
\end{cases} 
\\
y_r[i, j] &= \Bar{d}(p_a, p_j) ~~~~\text{if}~ y_o[i,j] = 1,
\label{eq:local_state}
\end{align}
where $p_a$, $p_j$ are the poses of the camera-wearer and object $o_j$ respectively, $d(p_a, p_j)$ is the euclidean distance between them, $\delta$ is a distance threshold for nearby objects (we set $\delta=5.0$m, beyond which visible objects are small), $\theta(p_a, p_j) \in [0...3]$ is the discretized angle of the object relative to the agent's heading (forward, right, behind, left), and $\Bar{d}(p_a, p_j) \in [0...4]$ is discretized distance. 
See \reffig{fig:local_state} and Supp.~for empirical analysis of related alternatives, e.g., predicting just object presence or image features.

We then train a model to predict the local state of the target video frame, conditioned on the video trajectory.
Once trained, the model can relate what is visible in a frame to the camera-wearer's possibly \emph{hidden} surroundings to produce environment-aware features. 

Since supervision for camera-wearer pose and every object's location is non-trivial for egocentric videos, %
where camera localization and tracking is error prone, we %
leverage videos in simulated environments for training, as presented next.

\vspace{-0.05in}
\subsection{Environment-aware pretraining in simulation} \label{sec:pretraining}

To source local state labels, we generate a dataset of video walkthroughs of agents in simulated 3D environments where agent and object poses are accessible at all times (see \refsec{sec:experiments}). We train our model in two stages described below.

\vspace{-0.05in}
\subsubsection{Pose embedding learning} \label{sec:pose_embed}
While ground truth camera pose is available from the simulator at training time, a model trained to rely on it will fail on real-world egocentric video at test time, where pose estimates are noisy.  %
With the goal of handling arbitrary indoor egocentric video, we instead explore representations that implicitly encode coarse pose information. See Supp. for pose-related experiments.

Specifically, for a sequence of RGB frames $\{f_t\}_{t=1}^T$ and camera poses $\{\theta_t\}_{t=1}^T$, we generate \emph{pose embeddings} $\{p_t\}_{t=1}^T = \mathcal{P}(f_1, ..., f_T)$ using a transformer encoder network. These embeddings are used to predict the relative pose between each observation pair using a bilinear layer
\begin{equation}
\hat{\theta}_{i,j} = p_i^T V_p p_j + W_p^T (p_j - p_i) + b_p,
\label{eq:pose_embed}
\end{equation}
where $\hat{\theta}_{i,j}$ is the predicted relative pose and $V_p, W_p, b_p$ are trainable parameters. %
We discretize the relative pose into $12$ angles and $4$ distance ranges to provide an approximate yet robust pose estimate. %
The network is trained to minimize cross-entropy between the predicted and the target relative pose labels for all observation pairs $\sum_{i,j} \mathcal{L}_{ce}(\hat{\theta}_{i,j}, \theta_{i,j})$. The trained pose embeddings $p_t$ encode information to help relate video observations based on their location and orientation in the environment.  %
Note that once trained, pose embeddings are inferred directly from video frame sequences --- ground truth pose is only required for training.

\vspace{-0.05in}
\subsubsection{Local state pretraining} \label{sec:local_state_pretraining}
Next, we train a model to embed visual information from a video walkthrough into an environment memory, which can then be queried to infer the local state corresponding to a given video frame from the same video. We implement this model as a transformer encoder-decoder model. 

Specifically, for a video walkthrough $\mathcal{V}$ with RGB frames $\{f_t\}_{t=1}^T$, and a query frame $f_q$,  we predict the local state $y_q = (y_o, y_r)$ as follows.
First, pose embeddings $\{p_t\}_{t=1}^T$ are generated for the video and query frames following \refsec{sec:pose_embed}. Then, each frame is encoded jointly with the pose embedding using a linear transform $\mathcal{M}_p$.
\begin{equation}
    x_t = \mathcal{M}_p([f_t; p_t)]). \label{eq:obs_encode}
\end{equation}

\begin{figure*}[t]
\centering
\includegraphics[width=\textwidth]{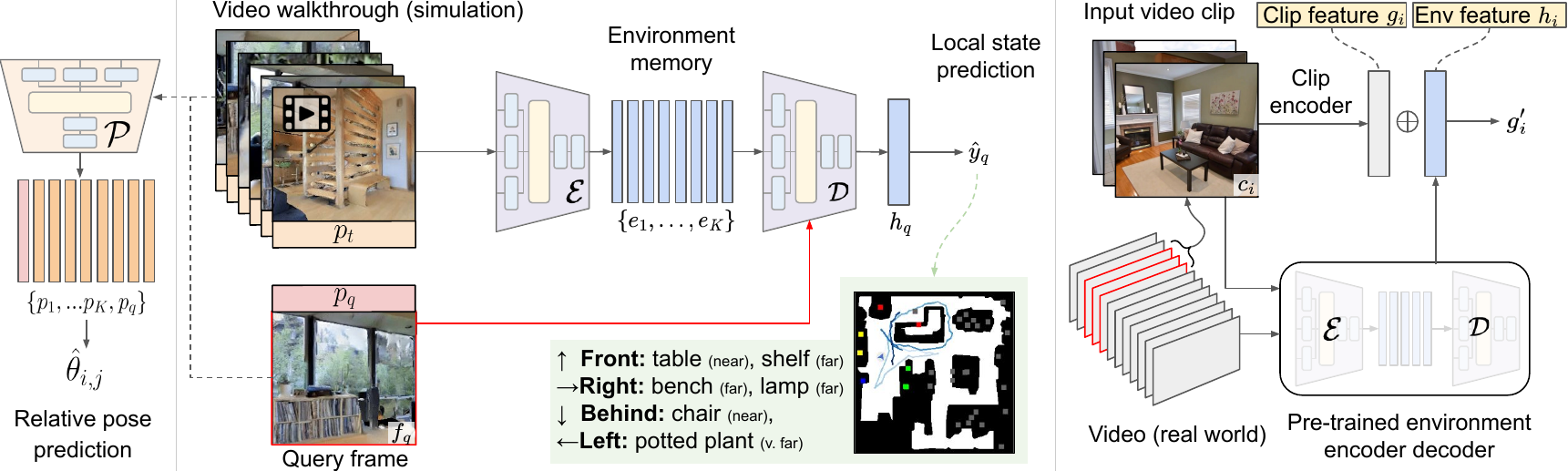}
\vspace*{-0.2in}
\caption{\textbf{Model framework.} \textbf{Left:} Our model first learns pose embeddings by predicting discretized relative pose between observations from simulated video walkthroughs (\refsec{sec:pose_embed}). \textbf{Center:} Next, it encodes observations and their pose embeddings into an \emph{environment memory} that is trained to predict the local environment state for a query frame (\refsec{sec:local_state_pretraining}). \textbf{Right:} Once trained, our model builds and queries the environment memory for any time-point of interest in a real-world video, to generate an environment feature for downstream video tasks in disjoint and novel scenes (\refsec{sec:downstream}). $\oplus$ = concatenation.} 
\label{fig:architecture}
\vspace*{-0.15in}
\end{figure*}

Next, we uniformly sample $K$ video frames to construct an environment memory using a transformer encoder $\mathcal{E}$, which updates frame representations using self-attention:
\begin{equation}
\left\{ e_1, ..., e_K \right\} = \mathcal{E}(x_1, ..., x_K). \label{eq:env_encode}
\end{equation}

The resulting memory represents features for each time-step that contain propagated information from all other time-steps. Compared to prior work~\cite{furnari2019would,zhang2020learning,lin2019bmn,liu2021context,zhao2021video}, our encoder has the ability to relate observations based on not just visual characteristics and their temporal ordering, but also their relative spatial layout in the environment. A transformer decoder $\mathcal{D}$ then attends over the memory using query $x_q$ to produce the output \Acronym~representation $h_q$:
\begin{equation}
h_q = \mathcal{D}(\left\{e_1, ..., e_K\right\}, x_q), \label{eq:env_decode}
\end{equation}
which is finally used to predict the local state using two linear classifiers $\mathcal{M}_o$ and $\mathcal{M}_r$ for object class and distance predictions respectively. The network is trained to minimize the combination of losses over the predicted and the target state labels for each direction:
\begin{equation*}
\mathcal{L}(h_q, y_o, y_r) = \mathcal{L}_{bce}(\mathcal{M}_o(h_q), y_o) + \lambda \mathcal{L}_{ce}(\mathcal{M}_r(h_q), y_r), \label{eq:loss}
\end{equation*}
where $\mathcal{L}_{bce}$ and $\mathcal{L}_{ce}$ refer to binary cross-entropy and cross-entropy losses respectively. We set $\lambda = 0.1$ which balances the contributions of each loss function based on validation experiments. The distance loss is computed only for objects that are in the local state ($y_o = 1$). See \reffig{fig:architecture} (left) and Supp.~for more architecture details. 

Learning to predict the local state involves aggregating information about observed  objects across time, as well as anticipating unseen objects based on learned priors from the layout of objects in other scenes (e.g., TVs are usually in front of couches; kitchens have particular arrangements of sinks, refrigerators, and stove-tops). 

Once trained, given a video in a new environment and a  time-point of interest, our model constructs an environment memory, predicting the local state based on information aggregated throughout the test video. 
Importantly, $h_q$ --- the \Acronym~feature --- contains valuable information about the camera-wearer's surroundings, offering environment features to complement traditional video features (e.g., for a person watching TV, also encode %
the couch they are sitting on, the lamp nearby).

\vspace{-0.05in}
\subsection{Environment-memory for video understanding with real videos} \label{sec:downstream}

\begin{figure*}[t]
\centering
\includegraphics[width=\textwidth]{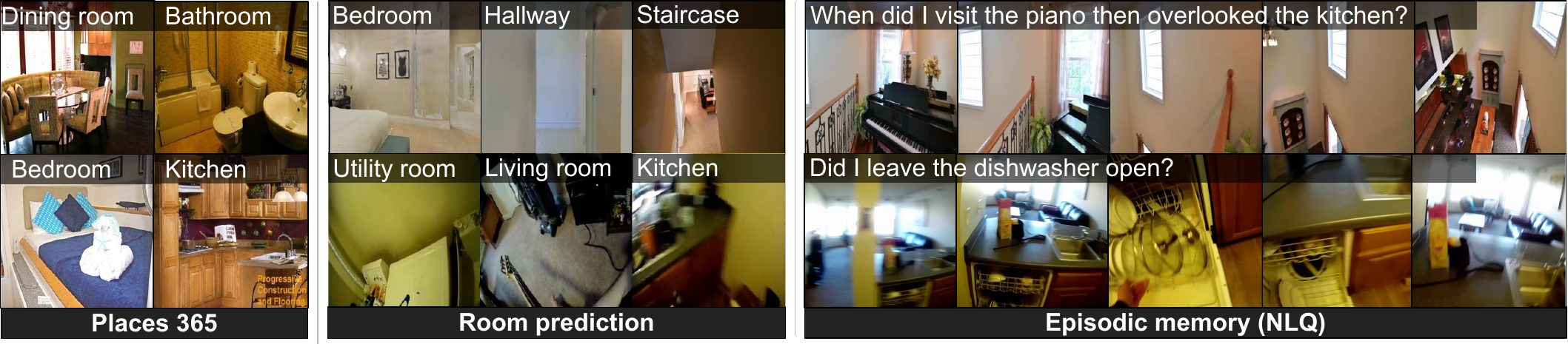}
\vspace*{-0.2in}
\caption{\textbf{Scene understanding in third-person photos vs. human-centric environment understanding.} \textbf{Left:} Well-framed, canonical images from Places365 are considerably different from scene content observed in egocentric video. \textbf{Center and right:} %
Real egocentric video streams from HouseTours (top) and Ego4D (bottom) illustrating %
the value in modeling %
the underlying environment, rather than just what is visible in short clips. For example, the person does not explicitly look at the staircase while walking down it (center, top row); the spatial relation between the person, piano, and kitchen is important to answer the question (right, top row).}
\vspace*{-0.15in}
\label{fig:task_example}
\end{figure*}

Next, we leverage our environment-memory model for real-world video tasks. A video understanding task defines a mapping from a sequence of video clips $\left\{c_1, ..., c_N\right\}$ from longer video $\mathcal{V}$ to a task label. 
We consider two tasks: (1) \B{\SC{RoomPred}}: where the model must classify which room $r_t$ the camera-wearer is in (e.g., living room, kitchen) at time $t$ in the video, and (2) \B{\SC{NLQ}}: natural language queries, an episodic memory task popularized recently in Ego4D~\cite{grauman2022ego4d} 
where the model must identify the temporal window $(t_{s}, t_{e})$ in the video that answers an environment-centric query $q$ specified in natural language. See \reffig{fig:task_example}. Both tasks entail human-centric spatial reasoning from video.

Current models produce clip features that encode only what is immediately visible. This is sufficient for short-horizon tasks (e.g., action recognition), but as we will show, falls short on the tasks above that require extra reasoning about the agent's surroundings. Our environment-memory model addresses this by enhancing standard clip features with context from the camera-wearer’s surroundings. 

To do this, for each input clip in $c_i \in \mathcal{V}$, we select the center frame $f_i$ of the clip as the query frame. Following \refsec{sec:pretraining}, we uniformly sample $K$ frames from the video around the query frame, encode them along with their pose embeddings (\refeqn{eq:obs_encode}), and build an environment memory using our environment encoder $\mathcal{E}$ (\refeqn{eq:env_encode}). Finally, we use our decoder $\mathcal{D}$ to produce the  output feature. This results in set of output \Acronym~features, one per input clip.
\begin{align}
    h_i & = \mathcal{D}(\left\{e^i_1, ..., e^i_K\right\}, x_i).
\end{align}
Each environment feature then enhances the original clip feature as follows
\begin{equation}
    g'_i = W_{\mathcal{E}}^T [g_i;h_i] + b_{\mathcal{E}}, \label{eqn:aggregate}
\end{equation}
where $g_i$ is the original clip feature for clip $c_i$ (e.g., ResNet, SlowFast, EgoVLP) and $W_{\mathcal{E}}, b_{\mathcal{E}}$ are linear transform parameters. See \reffig{fig:architecture} (right). 

The new clip features $\left\{g'_1, ..., g'_N\right\}$ consolidate features from what is directly visible in a short video clip and features of the (potentially unseen) space surrounding the camera-wearer. Put simply, our approach implicitly widens the field of view for tasks that reason about short video clips by providing a way to access features of their surroundings in a persistent, geometrically consistent manner.

Critically, we use the exact same \Acronym~representations to tackle both video understanding tasks.
This is a departure from traditional sim-to-real approaches where a task-specific dataset needs to be carefully designed for every downstream task, which may be impractical. %
That said, given that today's 3D assets (Matterport3D, HM3D, etc.) focus on indoor spaces, our %
model is best suited to videos in indoor environments. %
We discuss the sim-to-real gap in detail in Supp.  %

\section{Experiments} \label{sec:experiments}
We evaluate how our \Acronym~features learned in simulation benefit real-world video understanding. %

\vspace{-0.05in}
\paragraph{Simulator environments} For training, we use the Habitat simulator~\cite{savva2019habitat} with photo-realistic HM3D~\cite{ramakrishnan2021hm3d} scenes to generate simulated video walkthroughs. We generate $\aprox$15k walkthroughs from 900 HM3D scenes, each 512 steps long, taken by a shortest-path agent that navigates to randomly sampled goal locations %
(move forward, turn right/left $30^\circ$). For each time-step, we obtain the ground-truth local state from the simulator required in \refsec{sec:local_state} (i.e., object labels and relative pose). For object labels, we map instance predictions across $|\mathcal{O}| = 23$ categories to the 3D scenes using a pretrained instance segmentation model~\cite{fang2021instances} trained on the subset of scenes with semantic labels. Though the walkthroughs involve discrete actions, they share characteristics with real-world video (cameras at head-level; views covering the environment) making them %
suitable for transfer. See Supp. for examples and  details.   %

\vspace{-0.05in}
\paragraph{Video datasets} We evaluate our models on three egocentric video sources. 
(1) \textbf{HouseTours~\cite{chang2020semantic}} contains 119 hours of real-world video footage of house tours from YouTube. We use $\aprox$32 hours of video from 886 houses where the camera can be localized and create data splits based on houses.
(2) \textbf{Ego4D~\cite{grauman2022ego4d}} contains 3k hours of real-world video of people performing daily activities. We use all videos annotated for the NLQ benchmark and apply the provided data splits, which yields 1,259 unseen scenes.
(3) \textbf{Matterport3D (MP3D)~\cite{chang2017matterport3d}} contains simulated video walkthroughs from 90 photo-realistic 3D scenes. We use 146 long video walkthroughs and standard data splits~\cite{cartillier2021semantic}.

\begin{figure*}[t]
\centering
\includegraphics[width=\linewidth]{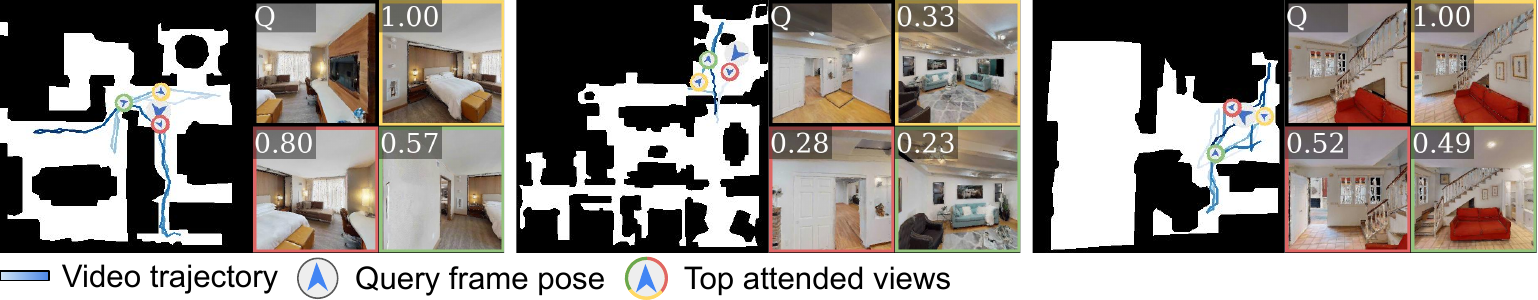}
\vspace{-0.2in}
\caption{\textbf{Visualized attention weights.} Our model learns to attend to views that help solve the local state prediction task. The query frame $Q$ and top-3 attended views (colored boxes), their positions along the trajectory (colored circles), and their associated attention scores are shown. See Supp. for more examples. 
}
\vspace{-0.2in}
\label{fig:attn_viz}
\end{figure*}

These datasets provide an ideal test-bed for our approach. %
On the one hand, both HouseTours and Ego4D are \emph{real-world} video datasets allowing us to test generalization to both real-world visuals as well as natural human activity across diverse, cluttered environments in unseen houses.
On the other hand, MP3D offers novel scenes with distinct visual characteristics and object distributions compared to HM3D, allowing us to test our model's robustness to domain shift in a controlled \emph{simulated} video setting. Note that none of the datasets have a 1-1 alignment in object taxonomy with HM3D, meaning our downstream tasks require generalization to both unseen environments and unseen objects.

We collect crowd-sourced labels for each task, which we will publicly share.
For \SC{RoomPred} these are room category labels from 21 classes (e.g., living room, kitchen) for each time-step on HouseTours and Ego4D. %
For \SC{NLQ} these are natural language queries and corresponding response tracks in the video. On HouseTours, we crowd-source queries (e.g., ``where did I last see my phone in the kitchen'', ``when did I first visit the bathtub'') and
on Ego4D, we use the official NLQ benchmark annotations, which require reasoning over actions, objects, and locations (e.g., ``what tool did I pick up from the table'', ``where did I hang the pink cloth'').
On MP3D, we source all labels directly from the simulator ($9$ room categories, and automatically generated NLQ queries from simulator object labels and locations). See Supp. for data collection details and \reffig{fig:task_example} for examples.

\vspace{-0.05in}
\paragraph{Experiment setup} 
For pre-training, we use $2048$-D ImageNet-pretrained ResNet50~\cite{he2016deep} features for each video frame. Our encoder-decoder models $\mathcal{P}, \mathcal{E}, \mathcal{D}$ are 2-layer transformers~\cite{vaswani2017attention} with hidden dimension $128$. $K=64$ frames are sampled from each video to populate the memory. We train models for 2.5k epochs and select the model with the lowest validation loss.
For \SC{RoomPred}, we generate a single \Acronym~feature aligned with the query clip. For \SC{NLQ} we generate one feature per input clip. 
Full architecture and training details are in Supp.

\vspace{-0.05in}
\paragraph{Baselines} For \SC{RoomPred} we use a popular scene recognition model \SC{PlacesCNN}~\cite{zhou2017places} as the baseline model. For \SC{NLQ} we use the state-of-the-art moment localization model \SC{VSLNet}~\cite{zhang2020span,lin2022egocentric,liu2022reler}. Within these two frameworks, we compare the following approaches to enhance clip representations: %
\B{\SC{FrameFeat}} uses a pretrained ResNet50~\cite{he2016deep} model to generate a frame feature corresponding to each clip. \B{\SC{ObjFeat}} trains an object detector on all available simulated HM3D data and generates backbone features for each clip. We use the QueryInst model~\cite{fang2021instances}. \B{\SC{MAE}~\cite{he2022masked}} trains a state-of-the-art self-supervised learning approach to reconstruct masked patches of walkthrough video frames. \B{\SC{EgoTopo}~\cite{nagarajan2020egotopo}} trains a graph convolutional network (GCN) over the video graph built following \cite{nagarajan2020egotopo}. %
\B{\SC{EPC}~\cite{ramakrishnan2021epc}} trains an environment memory model to predict masked \emph{zone} features conditioned on pose. \B{\SC{TRF (scratch)}} trains a scene-memory transformer model~\cite{fang2019scene} that shares our model architecture but is randomly initialized and fine-tuned for the task.

These baselines represent various strategies to incorporate environment information into clip representations ranging from frame features (\SC{MAE}, \SC{FrameFeat}, \SC{ObjFeat}), to topological graph-based features (\SC{EgoTopo}), to pose-based features (\SC{EPC}). Note that \SC{ObjFeat}, \SC{MAE} and \SC{EPC} all pre-train on the same walkthrough videos as our approach. \SC{ObjFeat} further benefits from ground-truth object labels from the simulator.
Features from these approaches augment the input clip representations following \refeqn{eqn:aggregate} --- baseline architectures remain unchanged. Note that \SC{EPC} requires privileged information---ground-truth camera poses at inference time---whereas our model does not.

\vspace{-0.05in}
\subsection{Pose embedding and local state pretraining} 
\label{sec:results_pretraining}
\vspace{-0.05in}
We begin by evaluating the pose embedding network trained to predict relative pose discretized into 12 angles and 4 distance ranges. On the validation set, the model achieves accuracies of 48.4\% on relative distance prediction and 34.4\% on relative orientation prediction. Note that this task is challenging --- models must predict relative pose for all possible pairs of observations in a trajectory using their visual features alone --- however the goal is to generate pose encodings, not to output perfect pose.
Next, we evaluate how well our model can infer the local state, reporting average precision (AP) in each direction. Given a \emph{forward} view, objects can be reliably recognized (37.8 AP) compared to naively outputting the distribution of objects seen at training (5.4 AP). Moreover, our approach can link views in the video trajectory to also infer and anticipate objects in other directions, i.e., to the right, left and behind (21.5, 24.9 and 20.2 AP respectively) given the forward view. %
We visualize the attention weights learned by our model to link relevant observations to the query in \reffig{fig:attn_viz}. Our model learns to select informative views for the task beyond just temporally adjacent frames or views with high visual overlap. For example, in the first image, the view with highest attention score (1.0) looks at the bed directly \emph{to the left} (yellow box), allowing our model to benefit from information beyond its field of view.

\vspace{-0.05in}
\subsection{\Acronym~features for room prediction} 
\label{sec:results_room_pred}
\vspace{-0.05in}
Next, we evaluate our method on predicting what room the camera-wearer visits in the video. All models have access to the full video, but inference is at at each time-step. %
The \SC{PlacesCNN} model is a 2-layer MLP classifier trained on features from a Places365~\cite{zhou2017places} scene classification model. 
Features are max-pooled across a clip of $N=8$ frames around the time-step of interest for additional context, %
as a single frame may be uninformative (e.g., facing a wall). We generate an environment feature aligned with the center of the clip.

\begin{figure*}[t]
\centering
\includegraphics[width=\textwidth]{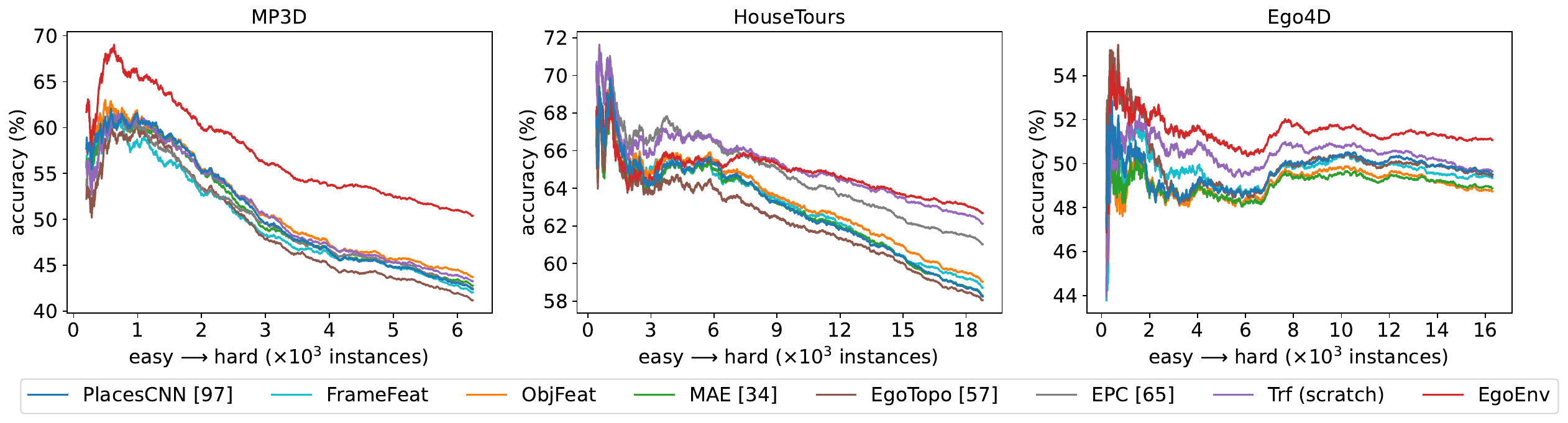}
\vspace{-0.20in}
\caption{\textbf{\SC{RoomPred} accuracy by instance difficulty.} Our method outperforms baselines, especially on \emph{hard} instances (smaller performance drop along curve). EPC requires pose at inference, which is unavailable in Ego4D. Results are aggregated across three runs. See Supp. for averaged results with error bars.
}
\vspace{-0.15in}
\label{fig:room_pred_results}
\end{figure*}

\reffig{fig:room_pred_results} shows the results. 
We report top-1 accuracy as a function of dataset difficulty, measured %
by the prediction entropy of the Places365 model trained on canonical scene images. Instances are ``hard'' (high entropy) where frame-level information is insufficient for predicting the room type.
Accuracy on the full dataset is at the far-right of the plots.
All models perform better on HouseTours compared to MP3D and Ego4D since the house tours were captured explicitly to provide informative views of each room. 
Despite having access to all additional pre-training videos and labels as supervision,  frame-level features from \SC{ObjFeat} and \SC{MAE} prove to be insufficient for environment-level reasoning.
All methods except ours perform worse
with the introduction of hard instances where the surrounding environment-context is important (left to right). This is a key result: despite training our models entirely in simulation, and with videos from a set of disjoint environments, our \Acronym~features are useful for downstream tasks on real-world videos.

\vspace{-0.05in}
\subsection{\Acronym~features for episodic memory}
\label{sec:results_nlq}
\vspace{-0.05in}
Next we evaluate on localizing the responses to natural language queries in egocentric video. 
We use \SC{VSLNet}~\cite{zhang2020span} and provide it with $N=128$ clips sampled uniformly from the full video to generate predictions. We use SlowFast~\cite{feichtenhofer2019slowfast} clip features and generate environment features aligned with each input clip. %
We use the benchmark-provided metric of \emph{Rank n@m}, which measures temporal localization accuracy~\cite{grauman2022ego4d}.

\reftbl{tbl:nlq_results} shows the NLQ results. Similar to \SC{RoomPred}, instances are harder in MP3D than in HouseTours as MP3D's shortest-path agents produce moments that are quick transitions between objects and locations, and contain only short glimpses of them. 
The global, video-level information from \SC{EgoTopo} improves performance slightly on all datasets. Strong image-level supervision (object labels in \SC{ObjFeat} and \SC{FrameFeat}) results in the largest improvements; \SC{MAE}, which has access to the same data but trains self-supervised representations, does not show strong improvements. \SC{EPC} %
performs well on MP3D,  where accurate pose is available from the simulator, but not on HouseTours with only noisy pose estimates (see Supp.). Our \Acronym~approach performs the best overall, outperforming even \SC{EPC}, which (unlike EgoEnv) has access to ground-truth pose at %
inference.

\begin{table*}[t]
\small
\centering
\begin{tabular}{|l|ccc|c|ccc|c|ccc|}
\multicolumn{1}{c}{}  &  \multicolumn{3}{c}{MP3D~\cite{chang2017matterport3d}} & \multicolumn{1}{c}{} & \multicolumn{3}{c}{HouseTours~\cite{chang2020semantic}} & \multicolumn{1}{c}{} & \multicolumn{3}{c}{Ego4D\textsuperscript{\textdaggerdbl}~\cite{grauman2022ego4d}} \\ 
\cline{1-4} \cline{6-8} \cline{10-12}
\SC{Rank1@m} $\rightarrow$  & @0.3 & @0.5 & \SC{avg} & & @0.3 & @0.5 & \SC{avg} & & @0.3 & @0.5 & \SC{avg} \\ 
\cline{1-4} \cline{6-8} \cline{10-12}
\SC{VSLNet}~\cite{zhang2020span}
                & 33.69     & 22.83     & 28.26     &
                & 42.94     & 27.68     & 35.31     & 
                & 5.45      & 3.12      & 4.29      \\
\SC{FrameFeat}  & 35.23     & 24.57     & 29.90     &
                & 48.45     & 32.06     & 40.25     & 
                & 5.58      & 3.28      & 4.43      \\
\SC{ObjFeat}    & 37.20     & 26.33     & 31.76     &
                & 47.74     & 32.49     & 40.11     & 
                & 5.76      & 3.43      & 4.59     \\
\SC{MAE}~\cite{he2022masked}
                & 35.11     & 24.13     & 29.62     &
                & 44.49     & 27.82     & 36.16     & 
                & 5.65      & 3.02      & 4.34      \\
\SC{EgoTopo}~\cite{nagarajan2020egotopo}
                & 36.10     & 25.06     & 30.58     &
                & 43.36     & 27.97     & 35.66     & 
                & 5.45      &  3.19     & 4.32       \\
\SC{EPC\textsuperscript{\textdagger}}~\cite{ramakrishnan2021epc}
                & 36.85     & \B{27.64} & 32.24     &
                & 43.22     & 27.82     & 35.52     & 
                & -         & -         & -        \\
\SC{Trf (scratch)}
                & 34.18     & 24.65     & 29.42     &
                & 40.54     & 22.32     & 31.43     & 
                & 5.25      & 3.12      & 4.19      \\
\SC{\Acronym}   & \B{38.18} & 26.85     & \B{32.52} & 
                & \B{51.98} & \B{34.18} & \B{43.08} & 
                & \B{6.04}  & \B{3.51}  & \B{4.77}  \\
\cline{1-4} \cline{6-8} \cline{10-12}
\end{tabular}
\vspace{-0.05in}
\caption{
\textbf{\SC{NLQ} results.} All baselines build on VSLNet~\cite{zhang2020learning} with alternate features. \textsuperscript{\textdagger}Privileged access to pose at inference, unavailable to our model, and absent in Ego4D. \textsuperscript{\textdaggerdbl}Validation split. See \reftbl{tbl:nlq_sota_results} for test set results. 
}
\label{tbl:nlq_results}
\end{table*}

\begin{table}[t]
\centering
\small
\begin{tabular}{|l|ccc|}
\hline
\SC{Rank 1@m} $\rightarrow$ & @0.3 & @0.5 & \SC{Avg} \\ \hline %
\SC{CONE}~\cite{hou2022efficient} & 15.26 & 9.24 & 12.25 \\ %
\SC{Badgers@UW-Mad.}~\cite{mo2022simple} & 15.71 & 9.57 & 12.64 \\ %
\SC{InternVideo}~\cite{chen2022internvideo} & 16.46 & 10.06 & 13.26 \\ %
\SC{NaQ}~\cite{ramakrishnan2023naq} & 21.70 & 13.64 & 17.67 \\ %
\SC{EgoEnv} & \B{23.28}	& \B{14.36}	& \B{18.82}	\\ \hline %
\end{tabular}
\vspace{0.05in}
\caption{\textbf{Ego4D \SC{NLQ} challenge results.} Our model obtains the best results against published approaches, and ranks 3rd among concurrent, unpublished work~\cite{hou2023groundnlq,shao2023action}.}
\vspace{-0.2in}
\label{tbl:nlq_sota_results}
\end{table}

Finally, \reftbl{tbl:nlq_sota_results} shows official eval server results for the Ego4D NLQ benchmark. We augment the baseline~\cite{ramakrishnan2023naq} with our \Acronym~features. We achieve state-of-the-art results, ranking 1st on the public leaderboard at the time of submission, and currently ranked 3rd.  %
Note that Ego4D contains in-the-wild videos of natural human activity in diverse scenes (e.g., workshops, gardens) compared to the simulated walkthrough videos in pretraining. Due to this \emph{sim-to-real} gap, our approach performs even better on instances aligned with training videos (navigation in indoor homes) as our Supp. experiments show. Our leading results on the full challenge set for this major benchmark demonstrates the value of our environment-centric feature learning approach, despite this gap.

\vspace{-0.05in}
\subsection{Analysis of sim-to-real gap} \label{sec:simtoreal}

Next, we discuss our approach in the context of the sim-to-real gap. Ego4D videos are in-the-wild, capture natural human actions and object-interactions, and take place in diverse scenes. These scenes may be significantly different from the simulated environments used for pre-training (navigation in indoor houses). In \reftbl{tbl:ego4d_nlq_breakdown}, we show results on the subset of videos that are aligned with the training environments on the Ego4D NLQ validation set. We select these using the scenario labels provided in Ego4D including indoor home scenarios (e.g., listening to music, household management) and navigation-heavy scenarios (walking indoors and outdoors) while excluding outdoor activities (e.g., golfing, outdoor cooking). See Supp. for the full list of scenarios. Our approach shows healthier improvements across both sets of scenarios, highlighting the effect of the sim-to-real gap. However, despite this gap, our approach is still able to outperform other approaches over all scenarios, demonstrating the value of our environment-centric feature learning approach.

\begin{table}[ht!]
\centering
\small
\begin{tabular}{|l|ccc|}
\hline
\SC{Rank k@m} $\rightarrow$ & R1@0.3 & R1@0.5 & \SC{avg} \\ \hline
\SC{NaQ (all)}~\cite{ramakrishnan2023naq} & 24.12 & 15.04 & 19.58  \\
\SC{EgoEnv (all)} & \B{25.37}	& \B{15.33}	& \B{20.35} \\ \hline
\SC{NaQ (indoor)}~\cite{ramakrishnan2023naq} & 28.91 & 17.97 & 23.44 \\
\SC{EgoEnv (indoor)} & \B{31.22}	& \B{19.09}	& \B{25.16}	\\ \hline
\SC{NaQ (nav)}~\cite{ramakrishnan2023naq} & 23.58 & 15.49 & 19.53 \\
\SC{EgoEnv (nav)} & \B{26.16}	& \B{16.01}	& \B{21.08}	\\ \hline
\end{tabular}
\vspace{0.05in}
\caption{\textbf{Ego4D \SC{NLQ} validation set results on aligned scenes.} Our approach performs better on the subset of videos that are aligned with our approach's simulated training environments (navigation, indoor houses).  
}
\vspace{-0.2in}
\label{tbl:ego4d_nlq_breakdown}
\end{table}

\vspace{-0.05in}
\subsection{Ablation experiments} \label{sec:ablations}
\vspace{-0.05in}
Next, we discuss some important ablations of our model design. We present full details of these experiments in Supp E and F, but we discuss the main conclusions here. 

\textbf{Importance of pose information:} We measure the effect of pose embeddings on our local state prediction task. Our models show small improvements in predicting objects in the forward view (+0.7 mAP) where scene information is directly visible, but large for other views that need to be inferred: mAP improvements of +2.2 (right), +0.9 (behind) and +1.0 (left). Further, we directly embed ground-truth pose as part of the input and see benefits on both tasks on MP3D, but not on HouseTours, due to noise in extracted pose (compared to simulator-provided pose in MP3D). 

\textbf{Alternate pretraining task formulations:} Next, we investigate alternate pretraining objectives instead of local state prediction including variants that only predict the object categories in each cardinal direction, but not the distances or that directly predicts the image features in each cardinal direction, among others. We find that our approach that requires predicting both object labels, orientations as well as rough distances offers a balance of both cues during pretraining, translating to strong downstream performance.

\textbf{Hyperparameter ablations:} We vary window size, memory size and the loss weighting term. We find that small window sizes are sufficient for localizing the room category for \SC{RoomPred}, while larger windows are required for \SC{NLQ}; and that the model is not very sensitive to the other parameters.

\section{Conclusion}
We proposed a framework to learn environment-aware representations in simulation and transfer them to video understanding tasks on challenging real-world datasets. Our approach outperforms state-of-the-art representations for predicting visited rooms and retrieving important moments from natural language queries, despite a significant sim-to-real gap. Despite its strengths, there are several opportunities for future work to improve our model further. These include incorporating 3D information into the local state task (currently defined in a 2D, top-down map), generating more \emph{human-like} simulated videos and integrating more explicit approaches to tackle the sim-to-real gap. 

\paragraph{Acknowledgements} Thanks to Fu-Jen Chu and Jiabo Hu for help collecting the HouseTours annotations. UT Austin is supported in part by IFML NSF AI institute. KG is paid as a research scientist at Meta.

{\small
\bibliographystyle{ieee_fullname}
\bibliography{egbib}
}

\clearpage

\maketitle

\pagenumbering{arabic}
\setcounter{page}{1}

\appendix

\setcounter{section}{0}
\setcounter{figure}{0}
\setcounter{table}{0}

\section*{Supplementary material}

This section contains supplementary material to support the main paper text. The contents include:

\begin{enumerate}[label*=, wide, labelindent=0pt, itemsep=0in]

    \item \ref{sec:supp_video}. \textbf{Videos illustrating our approach.}
    
    \item \ref{sec:ego4d_simtoreal}. \textbf{Discussion about the sim-to-real gap in Ego4D related to experiments in \refsec{sec:simtoreal}.}
    
    \item \ref{sec:supp_data}. \textbf{Data collection and annotation details.}
    \begin{enumerate}[label*=, itemsep=0in]
        \item \ref{sec:supp_walkthrough_details}. Additional walkthrough collection details to supplement \refsec{sec:experiments} (simulators). 
        \item \ref{sec:supp_data_collection_annotation}. Data annotation, processing details and analysis for both datasets in \refsec{sec:experiments} (video datasets).
    \end{enumerate}
    
    \item \ref{sec:supp_implementation_details}. \textbf{Implementation and training details.}
    \begin{enumerate}[label*=, itemsep=0in]
        \item \ref{sec:supp_pretrain_architecture_details}. Architecture and training details for all models in \refsec{sec:experiments} (baselines).
        \item \ref{sec:supp_downstream_architecture_details}. Architecture and training details for \SC{PlacesCNN} and \SC{VSLNet} in \refsec{sec:results_room_pred} and \refsec{sec:results_nlq}.
    \end{enumerate}
    
    \item \ref{sec:supp_experiments}. \textbf{Supplementary experiments and analysis.}
    \begin{enumerate}[label*=, itemsep=0in]
        \item \ref{sec:supp_mvp_results}. Experiments with extra pretraining data (Ego4D videos vs. simulator walkthroughs).
        \item \ref{sec:supp_pose_experiments}. Experiments with pose embeddings and ground truth pose inputs for fair comparisons with \SC{EPC} on both tasks.
        \item \ref{sec:supp_nlq_other_base}. EgoEnv integrated into other approaches besides \SC{VSLNet} in \reftbl{tbl:nlq_results}. 
        \item \ref{sec:supp_alternate_tasks}. Alternate local state task formulations compared to ours in \refsec{sec:local_state}. 
        \item \ref{sec:supp_mem_vs_ant}. Details about rare instances during pretraining following discussion in \refsec{sec:pretraining}.
        \item \ref{sec:supp_task_pretraining}. Experiments on task-specific pre-training in simulation following the discussion in \refsec{sec:downstream}.
    \end{enumerate}
    
    \item \ref{sec:supp_ablations}. \textbf{Ablations experiments and additional visualizations.}
    \begin{enumerate}[label*=, itemsep=0in]
        \item \ref{sec:supp_error_bars}. Results with error-bars corresponding to aggregate numbers in \refsec{sec:experiments}.
        \item \ref{sec:supp_extra_ablations}. Ablation experiments on model hyperparameters for our approach in \refsec{sec:pretraining} including loss weighting factor, memory size and window size.
        \item \ref{sec:supp_attn_viz}. Additional attention visualization results to supplement \reffig{fig:attn_viz}.
        \item \ref{sec:supp_easy_vs_hard}. Additional details about entropy and instance difficulty from \refsec{sec:results_room_pred}.
    \end{enumerate}
\end{enumerate}

\section{Videos illustrating our approach} \label{sec:supp_video}
We include our supplementary video on our project page \url{https://vision.cs.utexas.edu/projects/ego-env/}. The video illustrates the local state prediction task (\refsec{sec:pretraining}), downstream video tasks (\refsec{sec:downstream}) and our main results.

In the first part of the video, we demonstrate the local state prediction task from \refsec{sec:pretraining} of the main paper. The video shows the first-person view of the camera-wearer (left panel). The right panel shows the top-down view of the environment with the agent trajectory (blue gradient) and nearby objects (colored squares). Note that models only see the egocentric view --- the top-down map is for illustration only. Given a simulated video walkthrough and a query time-stamp, the model must predict the direction and rough distance of each object near it. Correct, missing and false positive predictions are shown for each direction (left panel). Correct predictions on the top-down map are highlighted in cyan. 

In the second part of the video, we show examples of the two downstream video tasks from \refsec{sec:downstream} of the main paper, on MP3D, HouseTours and Ego4D. In the \SC{RoomPred} examples, the model must predict which room the camera-wearer is in from a short clip. As mentioned in \refsec{sec:results_room_pred}, the clips show quick motions and often contain ambiguous views making it hard to predict room labels directly using traditional scene recognition methods. For example, the ``staircase'' is not visible as the person descends it (Ego4D, bottom right video).  In the \SC{NLQ} examples, the model must predict the moment in time that answers a particular environment-centric query. The video examples show how this requires reasoning about the camera-wearer's surroundings. For example, in the HouseTours clip (``when did I visit the sink in the bathroom?'') the sink is only seen briefly in the video but the response requires the window of time that the camera-wearer was physically near it (within arms reach) regardless of visibility. In the Ego4D clip (``Did I leave the drawer open''), the model must know where the drawer is relative to the camera-wearer and link their actions to this physical location in order to respond.

\section{Sim-to-real gap in Ego4D} \label{sec:ego4d_simtoreal}

As mentioned in \refsec{sec:simtoreal} of the main paper, our model is affected by the type and diversity of pretraining data --- videos of simulated agents walking around a house --- limiting its generalization to unconstrained real-world video. Similarly, our approach is limited by simulator functionality --- HM3D scenes support a small set of objects, which may not overlap with real-world environments, and Habitat does not support fine-grained object-interactions (e.g., chopping vegetables). As a result, we find that our approach works well on videos that are consistent with pretraining (i.e., indoor home scenarios; videos with lots of walking and less object interaction) but contributes less on out-of-distribution scenes and activities (e.g., golfing, outdoor cooking). The full list of scenarios is in \reftbl{tbl:aligned_scenarios}. 

\begin{figure*}[t]
\centering
\includegraphics[width=\textwidth]{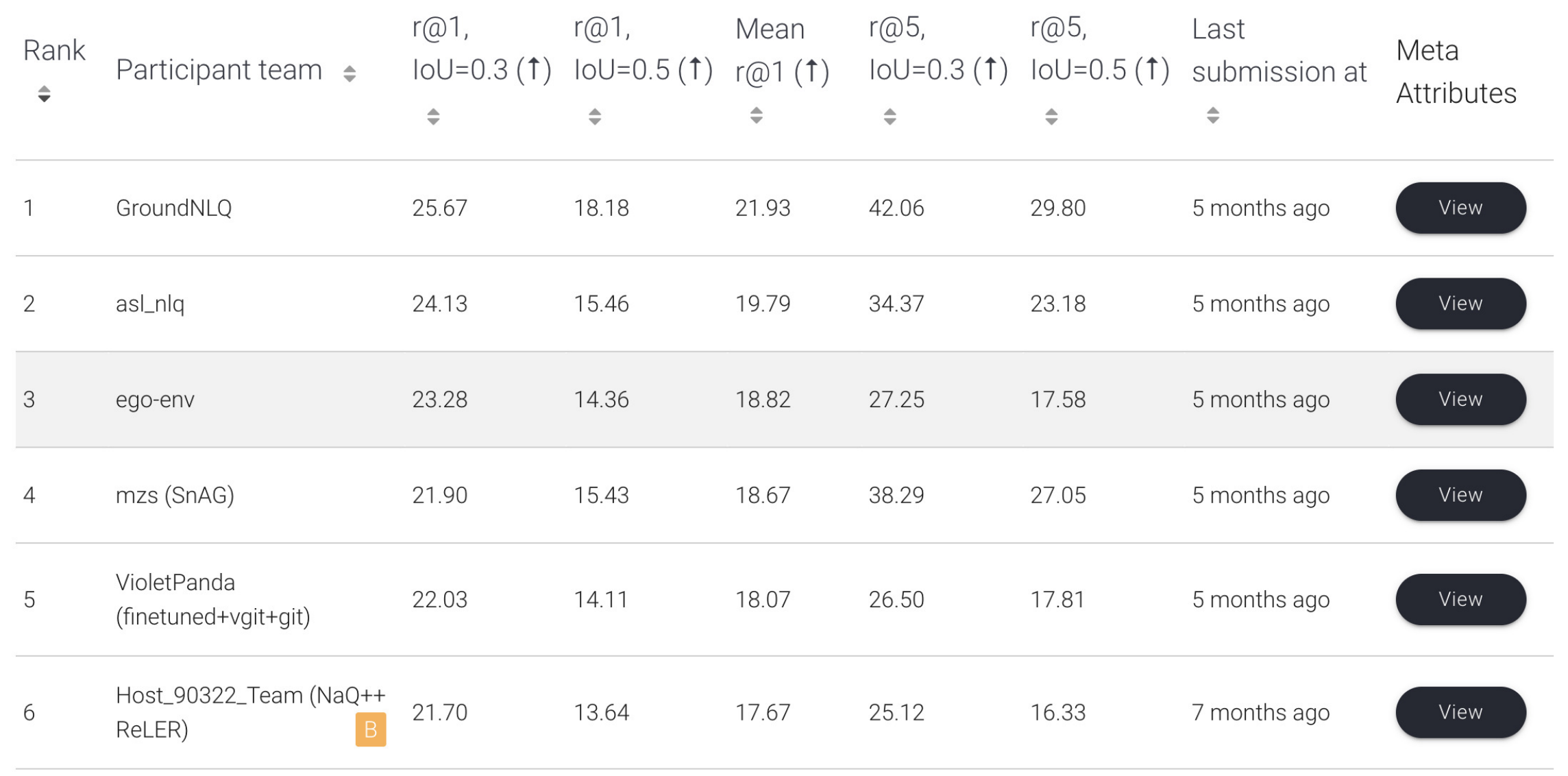}
\caption{\textbf{Snapshot of the Ego4D NLQ leaderboard on 10/26/23.} Leaderboard can be found on the \href{https://eval.ai/web/challenges/challenge-page/1629/leaderboard/3920}{challenge page}. Note that both GroundNLQ~\cite{hou2023groundnlq} and asl\_nlq~\cite{shao2023action} are concurrent works that were added to the leaderboard after the submission of this paper.} 
\label{fig:nlq_leaderboard}
\end{figure*}

Our results on the benchmark challenge (test set) in \reftbl{tbl:nlq_sota_results} corroborate this result. Note that our method was the top-ranked approach at the time of submission. Since then, other unpublished methods have been submitted to the leaderboard. We expect that future advancements in simulator capabilities (e.g., human motion models for agents, fine grained object interaction simulation) will help address this class of limitations. \reffig{fig:nlq_leaderboard} shows a snapshot of the leaderboard as of 10/23/23.

\begin{table}[ht!]
\centering
\small
\resizebox{\columnwidth}{!}{
\begin{tabular}{|l|p{5in}|}
\hline
Indoor & Visiting exhibition, On a screen (phone/laptop), Listening to music, Household management - caring for kids, Talking on the phone, Watching tv, Talking to colleagues, Electronics (hobbyist circuitry board kind, not electrical repair), Practicing a musical instrument, Eating, Cooking, Making coffee, Playing board games, Working at desk, Working out at home, Reading books, Playing games / video games, Hosting a party \\ \hline
Navigation & Roller skating, Bus, Walking on street, Indoor Navigation (walking), Walking the dog / pet, Car - commuting, road trip, Grocery shopping \\
\hline
\end{tabular}
}
\vspace{0.05in}
\caption{\B{List of scenarios aligned with training environments.}}
\vspace{-0.25in}
\label{tbl:aligned_scenarios}
\end{table}

\section{Data collection and annotation details.} \label{sec:supp_data}
We present data collection and annotation details for simulated walkthroughs, and our two downstream tasks (\SC{RoomPred} and \SC{NLQ}) for all three datasets (MP3D, HouseTours, Ego4D).

\subsection{Walkthrough generation details} \label{sec:supp_walkthrough_details}
\begin{figure*}[t]
\centering
\includegraphics[width=\textwidth]{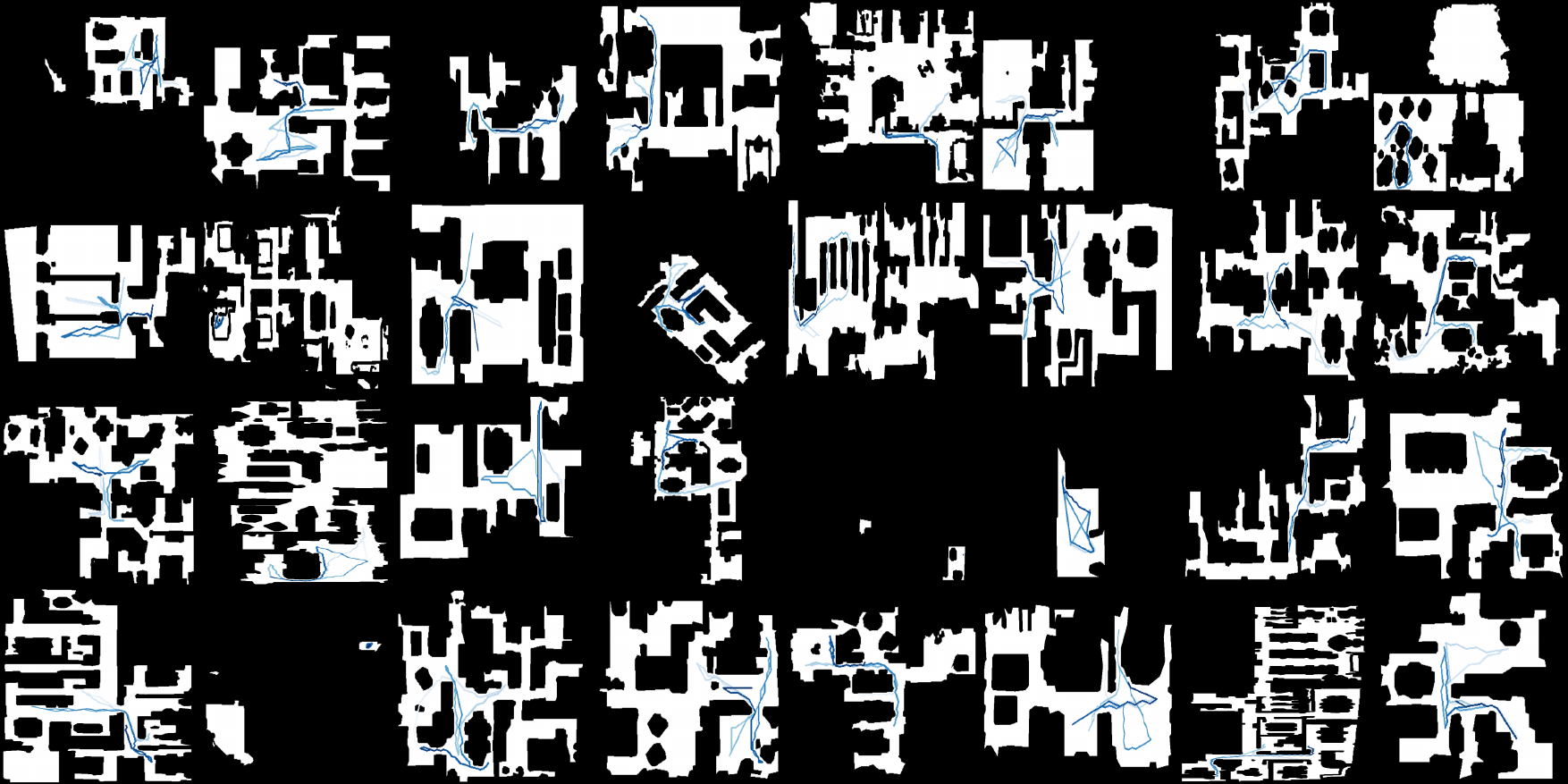}
\vspace*{-0.1in}
\caption{\textbf{Walkthrough examples in HM3D.} The blue gradient represents the trajectory from start (white) to end (blue). Black and white regions represent obstacles and free space respectively.
}
\label{fig:supp_walkthrough_examples}
\end{figure*}

As mentioned in \refsec{sec:experiments} (simulators) of the main paper, we generate simulated walkthroughs in HM3D~\cite{ramakrishnan2021hm3d} scenes to train our models. Given an environment, we first cluster all navigable points using KMeans, selecting between 4-64 clusters depending on the environment size. With each cluster centroid as a starting location, we sample 8-16 nearest goal locations, shuffle them (to allow re-visitation), and make an agent visit the goals in sequence. We use a shortest-path planning agent that uses the underlying navigation graph to reach goals in the fewest number of steps. We collect a dataset of $\aprox$15k episodes, each of 512 steps, of our agents visiting such goal sequences for experiments in \refsec{sec:pretraining}. \reffig{fig:supp_walkthrough_examples} shows a random sample of walkthroughs.

Note that the walkthroughs are generated in environments where objects are not moved, however a large part of real-world environments are in fact static. This includes static scene elements like doors, windows, counter tops, staircases, and most objects that are typically not moved like refrigerators, beds, couches, TV sets. Encoding these objects and scene elements can thus still provide value for human-centric environment understanding, even when some objects may have moved around.

\subsection{Data collection and annotation details} \label{sec:supp_data_collection_annotation}

\begin{table}[t]
\small
\centering
\begin{tabular}{|l|c|c|c|}
\hline
                                            & \SC{Scenes} & \SC{RoomPred}  &  \SC{NLQ}  \\ \hline
HM3D~\cite{ramakrishnan2021hm3d}            & 800 / 100 / --    & --                    &   --                  \\ 
Matterport3D~\cite{chang2017matterport3d}   & 57 / 6 / 21       & 1337 / 173 / 536    &   8380 / 837 / 3452  \\ 
HouseTours~\cite{chang2020semantic}         & 570 / 135 / 181   & 4512 / 1107 / 1593   &   2009 / 462 / 706     \\ 
Ego4D~\cite{grauman2022ego4d}                          & 998 / 328 / 333   & 3604 / 1596 / 1434   &   11291 / 3874 /4004     \\ \hline
\end{tabular}
\vspace{0.05in}
\caption{\textbf{Dataset train / val / test splits.} Splits based on scenes and instances per downstream task are shown. HM3D is used only for pretraining (\refsec{sec:pretraining}).}
\label{tbl:dataset_splits}
\vspace{-0.2in}
\end{table}

As mentioned in \refsec{sec:experiments} (simulators) of the main paper, we collect labels for each video dataset. For Matterport3D, we directly use the ground truth information available through the simulator to extract labels. For HouseTours and Ego4D, we crowd-source annotations where required. We describe the data collection process and present data statistics for each dataset and task. The resulting dataset splits can be seen in \reftbl{tbl:dataset_splits}.

\subsubsection{Annotation requirements} \label{sec:supp_annot_reqs}

\begin{table*}[t]
\resizebox{\textwidth}{!}{
\subcaptionbox*{}{
\centering
\begin{tabular}{|l|l|l|}
\multicolumn{3}{c}{\textbf{Matterport3D}} \\ \hline
hallway & bathroom      & bedroom               \\ \hline
office  & kitchen       & living room		    \\ \hline
lounge  & dining room   & family room           \\ \hline
\end{tabular}
}
\hfill
\subcaptionbox*{}{
\centering
\begin{tabular}{|l|l|l|l|l|}
\multicolumn{5}{c}{\textbf{HouseTours / Ego4D}} \\ \hline
attic               & balcony           & basement      & bathroom          & bedroom               \\ \hline
closet              & corridor/hallway  & dining room   & driveway          & front door/entrance   \\ \hline
garage/shed         & gym               & kitchen       & lawn/yard/garden  & living room           \\ \hline
office/home office  & porch             & \multicolumn{3}{l|}{recreation room (billiards room/play room)}   \\ \hline
staircase           & \multicolumn{2}{l|}{storage/laundry/utility room}      & swimming pool     &                       \\ \hline
\end{tabular}
}
}
\vspace{-0.2in}
\caption{\textbf{Room taxonomies for Matterport3D, HouseTours and Ego4D}}
\label{tbl:supp_room_taxonomy}
\end{table*}

\noindent \textbf{\SC{RoomPred}}
For this task, room labels are required at each time-step of the video. The $9$ room categories used in Matterport3D are in \reftbl{tbl:supp_room_taxonomy} (left). These categories are pre-defined in the simulator. The $21$ room categories used in HouseTours and Ego4D are in \reftbl{tbl:supp_room_taxonomy} (right). These categories were generated manually from a combination of Matterport3D room categories and a relevant subset of Places365 categories corresponding to indoor scenes.

\noindent \textbf{\SC{NLQ}}
For this task, natural language queries and corresponding moment boundaries (start and end times) are required. For HouseTours, we define $7$ query templates where $o$ refers to objects and $r$ refers to rooms: ``see $o$'', ``see $o$ in $r$'', ``see $o_1$ then $o_2$'', ``visit $r_1$ then $r_2$'', ``visit $o/r$'', ``visit $o_1$ then $o_2$'', ``visit $o$ in $r$''. Each template captures a type of question that requires a different mechanism of reasoning. ``see'' queries require reasoning about what is immediately visible; ``visit'' queries require an understanding of where the camera-wearer is in the environment and what objects are nearby (within arms reach); ``see/visit $o_1$ then $o_2$'' and ``see/visit $o$ in $r$'' require both spatial and temporal reasoning. Natural language queries follow from these templates. For example, for the ``visit $r_1$ then $r_2$'' template, a natural language query may be ``When did I first walk from the kitchen to the bathroom?''. The list of query templates, examples and descriptions can be seen in \reftbl{tbl:supp_nlq_query_templates}. The task definition follows prior work~\cite{grauman2022ego4d} but is adapted for the datasets used, and contains more environment-centric queries. For Ego4D, we use the existing NLQ benchmark task annotations.

Our video in \refsec{sec:supp_video} shows examples of both tasks to complement the image examples from \reffig{fig:task_example} of the main paper. The video highlights the stark contrast between prediction in static images (third-person photos) which contains well-framed images that are easy to recognize, and egocentric video which is much more challenging. In this setting, video is tied to quick ego-motion as the camera-wearer moves around the environment and objects are seen only briefly (or not at all) in non-canonical viewpoints. 

\begin{table*}[t]
\resizebox{\textwidth}{!}{
\begin{tabular}{|p{2.5cm}|p{6cm}|p{6cm}|}
\hline
\B{Template} & \B{Example} & \B{Description} \\ \hline
see $o$   &
Where did I first see the remote control? & 
Objects must be visible for the moment duration \\ \hline
see $o$ in $r$ &
When did I see the mirror in the bathroom? & 
Object must be physically inside the room \\ \hline
see $o_1$ then $o_2$ &
Where did I see a table then a chair?  & 
Objects can either be seen together or in quick succession \\ \hline
visit $r_1$ then $r_2$  &
When did I walk from the living room to the kitchen? &
Start = when the person begins to leave; End = they are fully inside the kitchen. \\ \hline
visit $o/r$             &
When did I last visit the couch?; When did I last visit the bedroom? &
Visit = physically near an object (within arms reach) or physically inside a room \\ \hline
visit $o_1$ then $o_2$  &
When did I visit the lamp then the couch? &
Same as see $o_1$ then $o_2$, but using the ``visit'' criteria above \\ \hline
visit $o$ in $r$        &
When did I visit the mirror in the bathroom? &
Same as see $o_1$ in $o_2$, but using the ``visit'' criteria above \\
\hline
\end{tabular}
}
\vspace{-0.05in}
\caption{\textbf{Query templates and examples for the \SC{NLQ} task on MP3D and HouseTours.}}
\label{tbl:supp_nlq_query_templates}
\end{table*}

\subsubsection{Matterport3D annotations}
For \SC{RoomPred}, navigable location are mapped to room categories using information from the simulator. We map agent positions for each video frame to these categories. For \SC{NLQ} we use room labels and extracted object positions to generate queries from the $7$ templates above. We define objects as ``seen'' if they occupy at least 5\% of the pixels in a given frame. We define objects as ``visited'' if the agent is $<$ 1.0m from the object, regardless of its visibility, following embodied navigation protocol (ObjectNav~\cite{batra2020objectnav}). Rooms are ``visited'' using the position $\rightarrow$ room category mapping above. We generate queries for each template by tracking objects and rooms that are seen and visited over time. If there are multiple visitations to an object or room, we ensure unique responses to queries by adapting them to consider only the first (or last) visit.

\begin{figure*}[t]
\centering
\includegraphics[width=\textwidth]{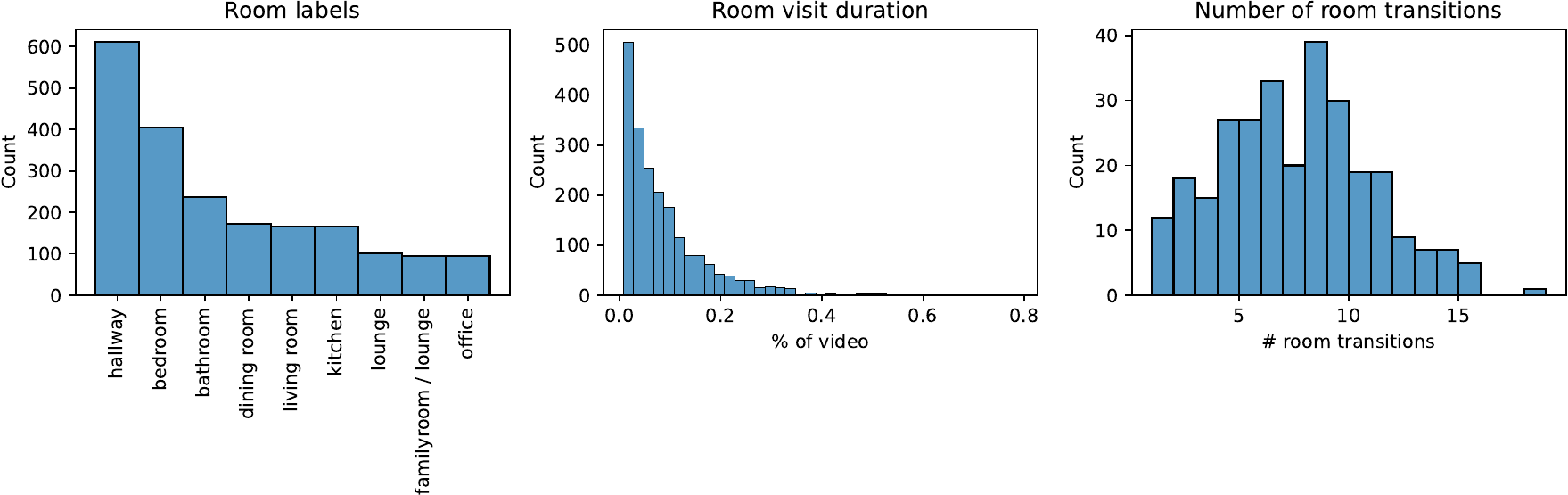}
\includegraphics[width=\textwidth]{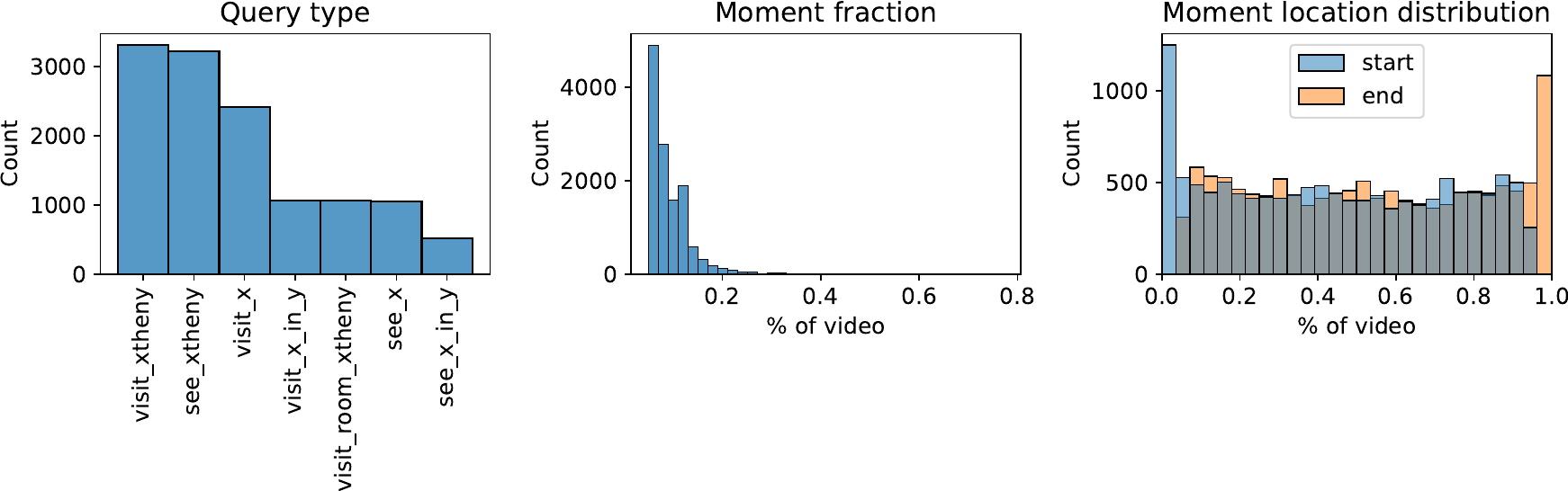}
\caption{\textbf{Annotation statistics for Matterport3D.} \textbf{Top panel:} \SC{RoomPred} data distribution. (left) distribution of room categories; (center) length of each room visit relative to the full video; (right) number of room transitions in each video. 
\textbf{Bottom panel:} \SC{Episodic memory retrieval} data distribution. (left) distribution of query types; (center) length of each annotated moment relative to the full video; (right) distribution of start and end times of annotated moments.
} 
\label{fig:supp_mp3d_annots}
\end{figure*}

\reffig{fig:supp_mp3d_annots} shows annotation statistics for both tasks on Matterport3D. Trajectories are fixed length (512 steps) with 7 room transitions on average. Both visits and moments are short, making it difficult to localize the response to the natural language query, and providing little extra context to recognize rooms. See our video in \refsec{sec:supp_video} for examples.

\begin{figure*}[t]
\centering
\includegraphics[width=\textwidth]{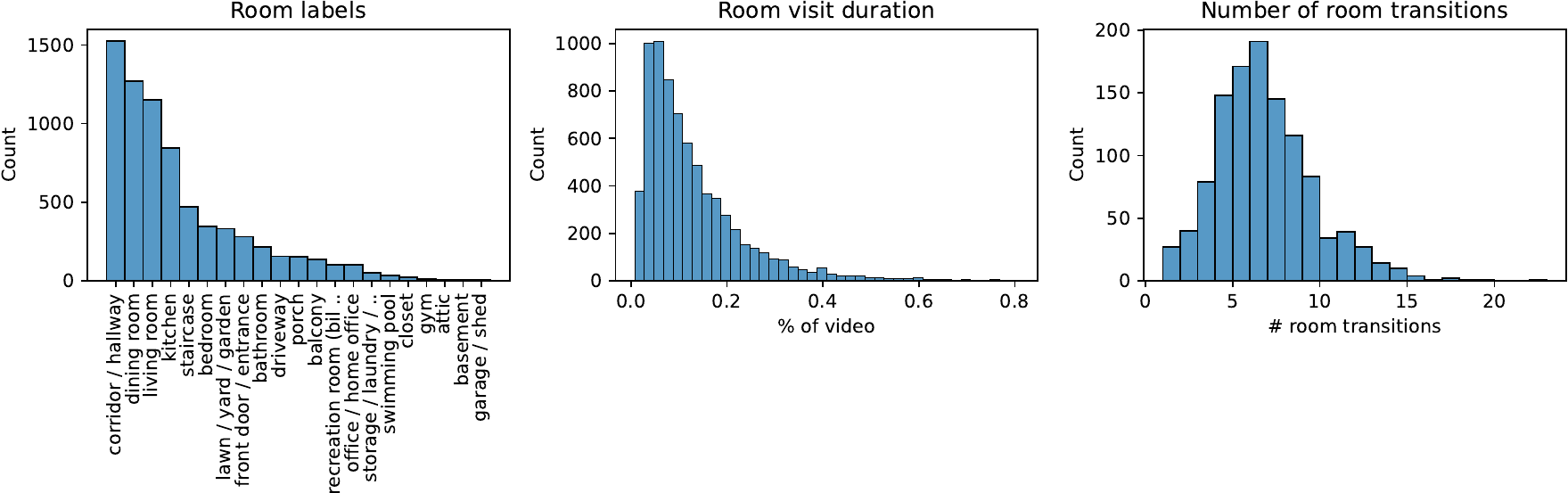}
\includegraphics[width=\textwidth]{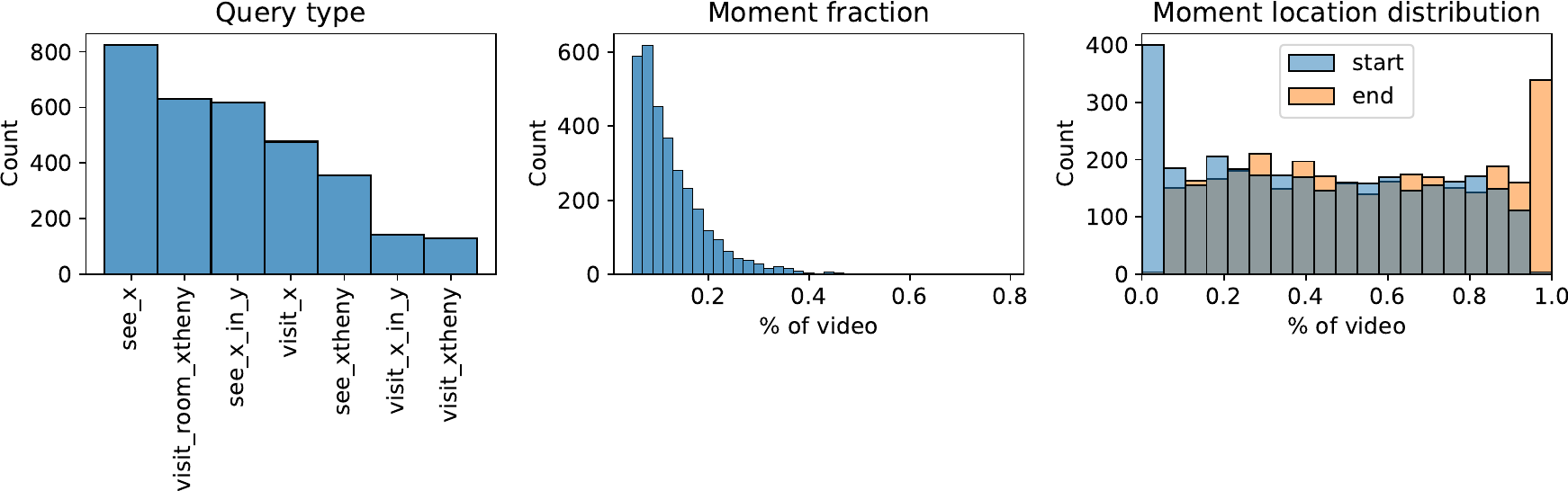}
\caption{\textbf{Annotation statistics for HouseTours.} \textbf{Top panel:} \SC{RoomPred} data distribution. (left) distribution of room categories; (center) length of each room visit relative to the full video; (right) number of room transitions in each video. 
\textbf{Bottom panel:} \SC{Episodic memory retrieval} data distribution. (left) distribution of query types; (center) length of each annotated moment relative to the full video; (right) distribution of start and end times of annotated moments.
} 
\label{fig:supp_housetours_annots}
\end{figure*}

\begin{figure*}[t]
\centering
\includegraphics[width=\textwidth]{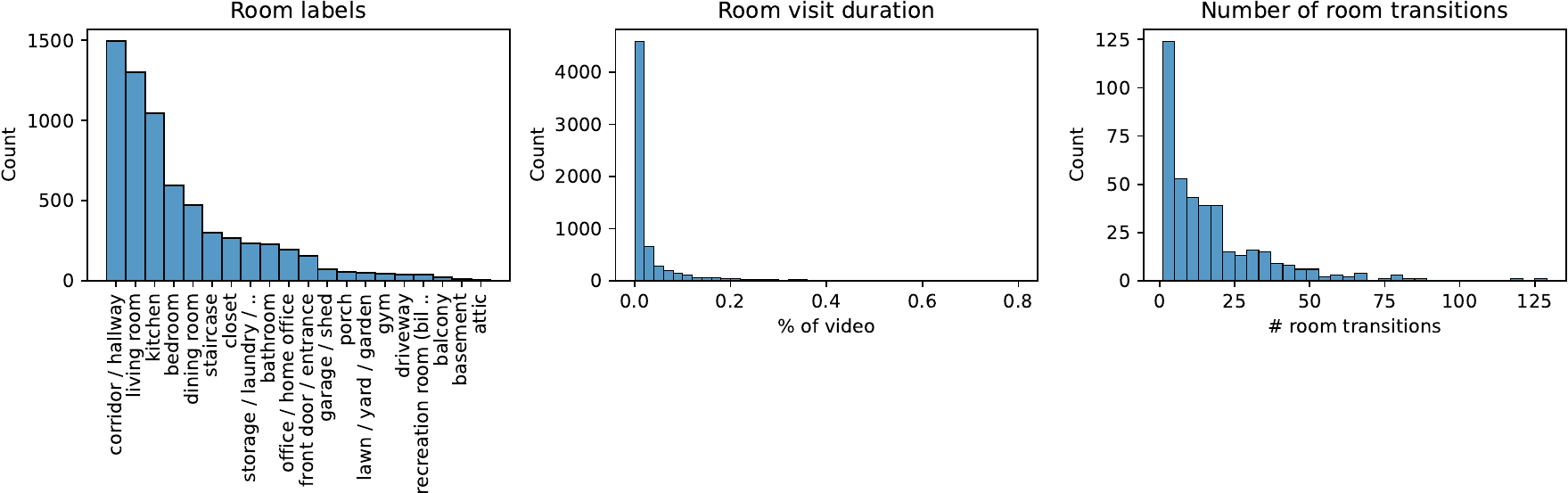}
\caption{\textbf{Annotation statistics for Ego4D.} \SC{RoomPred} data distribution. (left) distribution of room categories; (center) length of each room visit relative to the full video; (right) number of room transitions in each video. We do not collect annotations for \SC{NLQ} on Ego4D. We use annotations from the official Ego4D benchmark challenge (Section F.4. in the Ego4D paper~\cite{grauman2022ego4d}).} 
\label{fig:supp_ego4d_annots}
\end{figure*}

\subsubsection{HouseTours annotations}
\begin{figure*}
\begin{minipage}[c]{\textwidth}
\includegraphics[width=\textwidth]{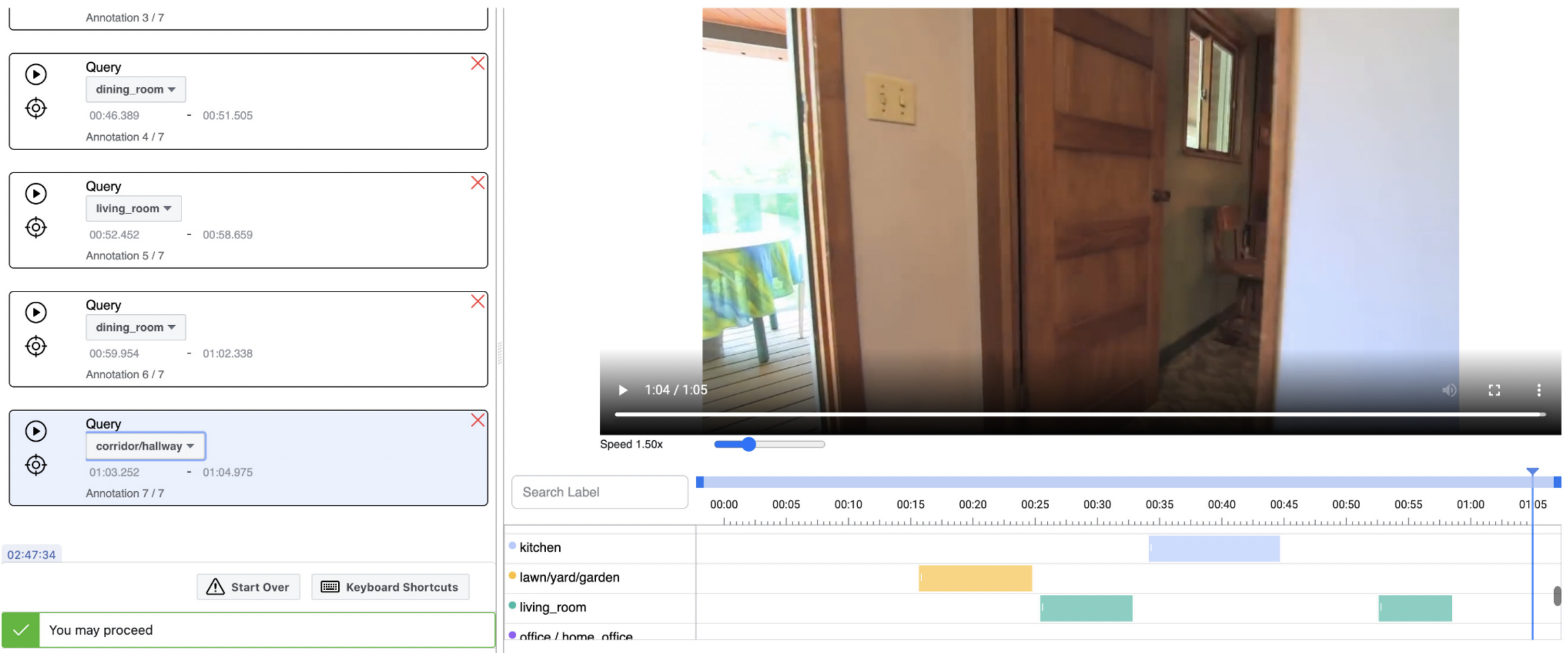}
\caption{\textbf{Annotation interface for collecting \SC{RoomPred} labels.} Annotators must densely segment room visits (start and end times) and associate a class label to each of them.} 
\label{fig:supp_roompred_interface}
\end{minipage}
\begin{minipage}[c]{\textwidth}
\includegraphics[width=\textwidth]{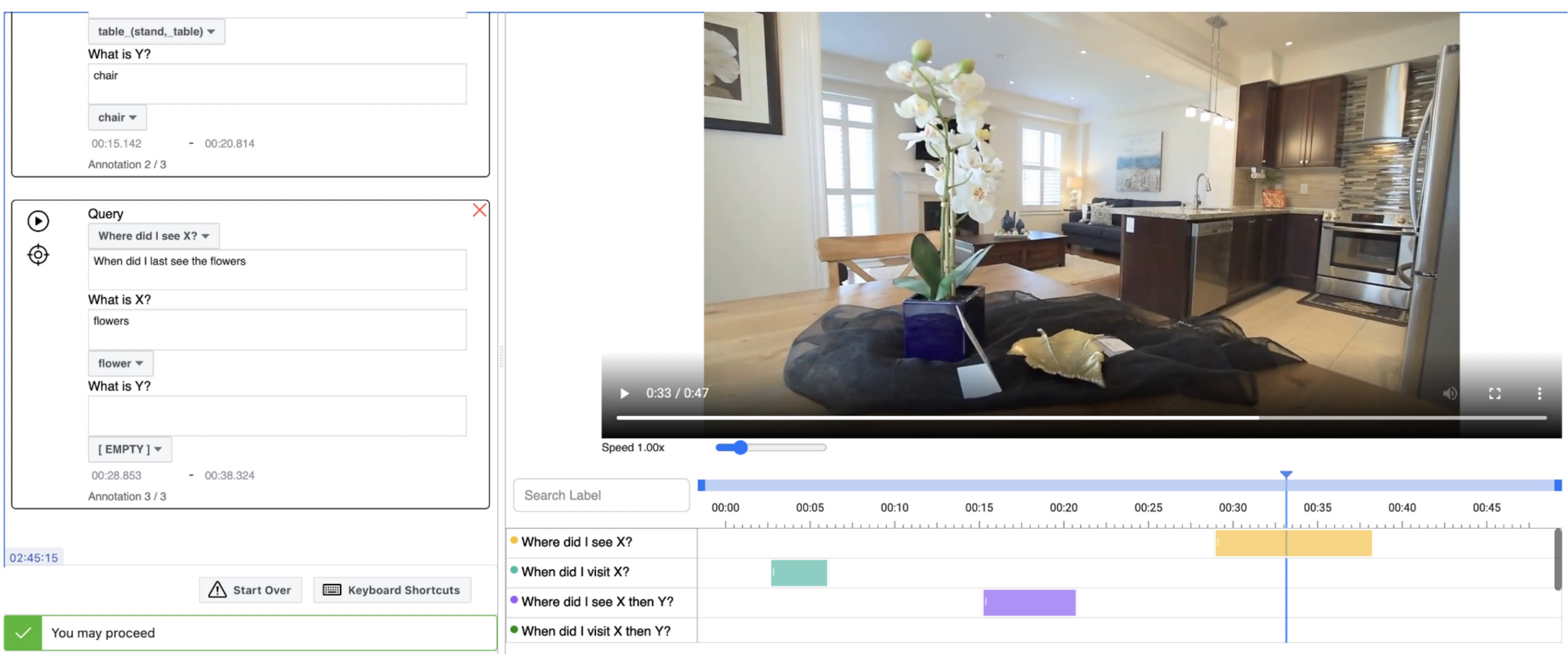}
\caption{\textbf{Annotation interface for collecting \SC{NLQ} labels.} Annotators must identify an interesting moment in time (start and end time) and associate a query template, a natural language query and object/room labels that fill the query template slots.} 
\label{fig:supp_nlq_interface}
\end{minipage}
\end{figure*}

We crowd-source annotations for real-world videos from HouseTours. For \SC{RoomPred}, we ask annotators to watch a video and mark the start and end time of each ``visit'' to a room. They must then label each visit with one of the $21$ room categories (from \reftbl{tbl:supp_room_taxonomy}, right). An illustration of the annotation interface is shown in \reffig{fig:supp_roompred_interface}.

For \SC{NLQ}, we ask annotators to identify an interesting moment (e.g., where a person sees a salient object, moves from one room to another, visits an important object) which serves as the answer to a query. The moment is specified by a start and end time, while the query is specified as natural language text generated following one of the $7$ template classes from \reftbl{tbl:supp_nlq_query_templates}. An illustration of the annotation interface is shown in \reffig{fig:supp_nlq_interface}.

\reffig{fig:supp_housetours_annots} shows annotation statistics for both tasks on HouseTours. In general, trajectories are relatively shorter than Matterport3D, though they involve a similar number of room transitions (6 on average). They share similar challenges with short moments. See our video in \refsec{sec:supp_video} for examples.

\subsubsection{Ego4D annotations}
Following the same procedure as HouseTours, we crowd-source room visit labels on Ego4D videos (see \reffig{fig:supp_ego4d_annots}). Ego4D videos are much longer (30 mins on average). As a consequence, visits are naturally a much smaller fraction of the overall video and the number of room transitions are much higher (16 on average). The distribution of room categories is similar to HouseTours. Annotation statistics for NLQ can be found in Section F.4 of the supplementary material of the Ego4D paper~\cite{grauman2022ego4d}.

\clearpage
\clearpage

\section{Implementation and training details} \label{sec:supp_implementation_details}
We present architecture and training details for all approaches.

\subsection{Architecture and training details (pretraining)} \label{sec:supp_pretrain_architecture_details}
We present additional details for our model in \refsec{sec:approach} of the main paper, as well as the baselines in \refsec{sec:experiments} (baselines). The full list of hyperparameters are in \reftbl{tbl:supp_pretrain_hyperparams}.

\vspace{0.05in}
\noindent \textbf{Pose embedder $\mathcal{P}$, environment encoder $\mathcal{E}$ and decoder $\mathcal{D}$.}
We build on the transformer~\cite{vaswani2017attention} architecture for our model. The inputs $f \in \mathbb{R}^{2048}$ are features from an ImageNet-pretrained ResNet-50 . These are encoded into 128-dimension vectors using a visual encoder (2-layer MLP). 

First, 128-$D$ pose embeddings are generated following \refeqn{eq:pose_embed}. These pose embeddings are concatenated with the visual embedding, and then transformed back to a 128-dimension vector using $\mathcal{M}_p$. Sin-cosine position embeddings are added to this input following \cite{vaswani2017attention} resulting in the transformer input $\{x_1, ..., x_N\}$ in \refeqn{eq:obs_encode}. Note that a separate visual encoder generates input features for the encoder $\mathcal{E}$ and the pose embedder $\mathcal{P}$.

The encoder $\mathcal{E}$ performs multi-headed self-attention using these inputs, with 2 layers, 8 heads and hidden dimension $128$. The decoder $\mathcal{D}$ is a 2-layer transformer with hidden dimension $128$ which attends to the outputs of $\mathcal{E}$ to generate the output representation. We predict relative directions for 23 object classes in HM3D (see \reftbl{tbl:supp_object_taxonomy}). We use the transformer implementation from PyTorch~\cite{paszke2019pytorch}. 

Our architecture is similar to prior embodied navigation approaches~\cite{ramakrishnan2021epc,fang2019scene} but does not require pose (and instead uses pose embeddings), includes the environment decoder that queries the memory based on pose embeddings and visual content, and is trained using our proposed learning objective in \refsec{sec:pretraining}. See \reftbl{tbl:supp_pretrain_hyperparams} (left).

\begin{table}[t]
\small
\centering
\begin{tabular}{|l|l|l|l|}
\hline
chair	    & table	        & picture	& cabinet	            \\ \hline
cushion	    & sofa	        & bed	    & chest\_of\_drawers	\\ \hline
plant	    & sink	        & toilet	& stool	                \\ \hline
towel	    & tv\_monitor	& shower	& bathtub	            \\ \hline
counter	    & fireplace	    & shelving	& seating	            \\ \hline
furniture	& appliances	& clothes   &	                    \\ \hline
\end{tabular}
\caption{\textbf{HM3D object taxonomy}}
\label{tbl:supp_object_taxonomy}
\end{table}

\vspace{0.05in}
\noindent \textbf{EPC architecture details.} 
EPC~\cite{ramakrishnan2021epc} is a transformer encoder-decoder model that masks out frames from physical locations and predicts the features of these masked zones given a query pose. We generate graphs to train this baseline following their approach. 

Specifically, for every video frame $\{f_1, ..., f_T\}$, we compute the geometric viewpoint overlap with every other frame. The viewpoint overlap $\psi(f_i, f_j)$ is calculated by projecting pixels from the frames to 3D point-clouds using camera intrinsics, agent pose and depth measurements, and measuring the percentage of shared points across frames. We use $\psi(f_i, f_j)$ as our distance metric to cluster all frames in video into zones using hierarchical agglomerative clustering. We set the distance threshold to $0.8$ as larger values result in too few zones.

We sample 4 unseen zones per instance in a batch of which one is ``positive'' and three are ``negatives'' for contrastive learning. We collect negatives from all instances in a batch during training. The network is trained using noise-contrastive estimation following \cite{ramakrishnan2021epc}. See \reftbl{tbl:supp_pretrain_hyperparams} (center).

\vspace{0.05in}
\noindent \textbf{EgoTopo architecture details.} 
EgoTopo~\cite{nagarajan2020egotopo} is an approach to translate egocentric video frames into a topological graph, where each node contains a list of clips that correspond to a physical location, and edges correspond to rough spatial layout.

For MP3D and HouseTours, ee directly use available pose information to determine whether two frames belong to the same zone or not (i.e., we do not train a retrieval network to approximate this). This amounts to a clustering of the trajectory based on pose (position and heading) and represents an enhanced version of EgoTopo that benefits from pose data. We use affinity propagation to cluster frames into nodes. To compute edges, we calculate the distance between the centroid of each node and assign an edge if this distance is $<$ 3.0m to be consistent with \refeqn{eq:local_state}. For Ego4D, pose information is not available. We fall back to clustering visual frames based on ImageNet pre-trained ResNet-50 frame features. 

Node features are calculated as the average of features assigned to that node. We use a 2-layer graph convolutional neural network (GCN) to aggregate features across nodes, and then average them to form a single video encoding following \cite{nagarajan2020egotopo}. See \reftbl{tbl:supp_pretrain_hyperparams} (center).

\vspace{0.05in}
\noindent \textbf{MAE architecture details.} 
We use all frames from our generated walkthroughs to train an MAE VIT-large model with $16\times 16$ patch size. We use the authors \href{https://github.com/facebookresearch/mae/blob/main/PRETRAIN.md}{existing code} and train a model for 200 epochs. 

\vspace{0.05in}
\noindent \textbf{ObjFeat architecture details.} We train a QueryInst instance segmentation model~\cite{fang2021instances} with a R-50-FPN backbone, using a dataset sampled from the 100 HM3D scenes with ground-truth semantic object annotations. We use the implementation in the \href{https://github.com/open-mmlab/mmdetection/tree/master/configs/queryinst }{MMDetection package}.

\vspace{0.05in}
\noindent \textbf{Training details.} 
As mentioned in \refsec{sec:experiments} (experiment setup) of the main paper, we train our models for 2500 epochs using the Adam optimizer with learning rate $1e-4$.  We sample $K=64$ frames to construct our memory. At training, we sample frames from the video but randomly offset frame indices to train robust models. %
During inference, we uniformly sample frames. We select the model with the lowest validation loss to evaluate downstream. See \reftbl{tbl:supp_pretrain_hyperparams} (right) for all optimization hyperparameters.

\begin{table*}[t]
\centering
\small
\hfill
\subcaptionbox*{}{
\centering
\begin{tabular}{lc}
\toprule
\multicolumn{2}{c}{Transformer architecture}                 \\ \midrule
Input dim                              & 2048                \\
Hidden size                            & 128                 \\
Pose emb dim                           & 128                  \\
Walkthrough length                     & 512                 \\
Memory size (K)                        & 64                  \\
$\#$ attention heads                   & 8                   \\
$\#$ encoder layers                    & 2                   \\
$\#$ decoder layers                    & 2                   \\
\bottomrule
\end{tabular}
}
\hfill
\subcaptionbox*{}{
\centering
\begin{tabular}{lc}
\toprule
\multicolumn{2}{c}{EPC parameters}                           \\ \midrule
$\#$ Unseen zones                      & 4                   \\
Clustering threshold                   & 0.8                 \\ \midrule
\multicolumn{2}{c}{EgoTopo parameters}                      \\ \midrule
Affinity prop damping                  & 0.5-0.9             \\ 
GCN hidden dim                         & 128                 \\ 
$\#$ GCN layers                        & 2                   \\ 
\bottomrule
\end{tabular}
}
\hfill
\subcaptionbox*{}{
\centering
\begin{tabular}{lc}
\toprule
\multicolumn{2}{c}{Optimization parameters}                  \\ \midrule
Max epochs                             & 2500                 \\
Learning rate                          & 1e-4                \\
Weight decay                           & 2e-5                \\
Batch size                             & 512                \\
\bottomrule
\end{tabular}
}
\hfill
\caption{\textbf{Architecture and training hyperparameters for pretraining (\refsec{sec:pretraining})}}
\label{tbl:supp_pretrain_hyperparams}
\end{table*}

\subsection{Architecture and training details (downstream)} \label{sec:supp_downstream_architecture_details}

We present additional details for the approaches in \refsec{sec:downstream} of the main paper. The full list of hyperparameters are in \reftbl{tbl:supp_downstream_hyperparams}.

\vspace{0.05in}
\noindent \textbf{Room prediction models.} 
As mentioned in \refsec{sec:results_room_pred} of the main paper, for our \SC{PlacesCNN} baseline model, we build a classifier on top of features from the wide ResNet-18 model from \cite{zhou2017places}. We use the authors \href{https://github.com/CSAILVision/places365}{existing code} and their provided pretrained models to initialize the model. The classifier head is a 2-layer MLP with hidden dimension $512$. The backbone is frozen and only the classifier is fine-tuned. $N=8$ frames around the target frame are used to provide additional context. The features are max-pooled before classification.

For our models, we use the target frame as the query to produce a single environment-feature, which is then concatenated with all $N=8$ frames and aggregated following \refeqn{eqn:aggregate}. This new, enhanced input is fed into the baseline model as described above.

We train all models for 80 epochs with learning rate $1e-4$ using the Adam optimizer. The full list of hyperparameters are in \reftbl{tbl:supp_downstream_hyperparams} (left)

\vspace{0.05in}
\noindent \textbf{Episodic memory retrieval models.} 
As mentioned in \refsec{sec:results_nlq} of the main paper,  we build on the VSLNet model from \cite{zhang2020span}. We use an \href{https://github.com/EGO4D/episodic-memory}{existing implementation} based on the authors original code. Visual inputs are encoded as $N=128$ clip features, created by adaptive average pooling of SlowFast-R50 clip features. Note that for MP3D we use ResNet-50 features as the walkthroughs contain discrete agent steps, rather than smooth video frames. 

For our models, we select the center frame of each of the $N=128$ inputs and use them as query frames to produce $N=128$ environment features. Each pair of input feature and environment feature is aggregated following \refeqn{eqn:aggregate}, and then input to the VSLNet model described above. Note that we perform this aggregation \emph{after} the video affine and feature encoding layers in VSLNet.

We train all models for 200 epochs with a learning rate of $1e-4$ using the Adam optimizer. The full list of hyperparameters are in \reftbl{tbl:supp_downstream_hyperparams} (right). 

For Ego4D experiments, we use the hyperparameters described in the respective benchmark whitepapers~\cite{grauman2022ego4d,lin2022egocentric,liu2022reler} and only aggregate precomputed environment features as described above.

\begin{table*}[t]
\centering
\small
\hfill
\subcaptionbox*{}{
\centering
\begin{tabular}{lc}
\toprule
\multicolumn{2}{c}{\SC{RoomPred}}                   \\ \midrule
$\#$ input frames                   & 8             \\
Hidden size                         & 512           \\
$\#$ layers                         & 2             \\ \midrule
\multicolumn{2}{c}{Optimization parameters}         \\ \midrule
Max epochs                          & 80            \\
Learning rate                       & 1e-4          \\
Weight decay                        & 2e-5          \\
Batch size                          & 512           \\
\bottomrule
\end{tabular}
}
\hfill
\subcaptionbox*{}{
\centering
\begin{tabular}{lc}
\toprule
\multicolumn{2}{c}{\SC{NLQ}}                        \\ \midrule
$\#$ Input clips                    & 128           \\
Hidden size                         & 128           \\
Highlight lambda                    & 5.0           \\
Extend boundary \%                  & 0.1           \\
$\#$ heads                          & 8             \\  
$\#$ layers                         & 4             \\  
Dropout rate                        & 0.2           \\ \midrule
\multicolumn{2}{c}{Optimization parameters}         \\ \midrule
Max epochs                          & 200           \\
Learning rate                       & 1e-4          \\
Weight decay                        & 1e-2          \\
Batch size                          & 64            \\
\bottomrule
\end{tabular}
}
\hfill
\hfill
\caption{\textbf{Architecture and training hyperparameters for \SC{RoomPred} and \SC{NLQ} (\refsec{sec:downstream})}}
\label{tbl:supp_downstream_hyperparams}
\end{table*}

\section{Supplementary experiments and analysis.} \label{sec:supp_experiments}

\subsection{Experiments with extra pretraining data (Ego4D videos vs. simulator walkthroughs)} \label{sec:supp_mvp_results}

In \refsec{sec:experiments} (baselines), all approaches have access to the same set of simulator-generated walkthrough videos for pre-training for apples-to-apples comparisons. In this section, we test the effect of directly performing self-supervised pretraining on Ego4D videos instead. In \reftbl{tbl:mvp} we compare our \textsc{MAE} baseline trained on walkthrough videos with the same architecture trained on Ego4D frames~\cite{xiao2022masked} using the author provided \href{https://github.com/ir413/mvp}{pretrained model}. 

\begin{table}[ht]
\centering
\scriptsize
\begin{tabular}{|l|ccc|}
\hline
\textsc{Rank1@m} $\rightarrow$ & @0.3 & @0.5 & \textsc{avg} \\ \hline
\textsc{MAE (Walkthrough)}   & 5.65 & 3.02 & 4.34 \\ 
\textsc{MAE (Ego4D)}         & 5.65 & 3.27 & 4.47 \\ 
\textsc{EgoEnv}  & \textbf{6.04}  & \textbf{3.51} & \textbf{4.77} \\ \hline
\end{tabular}
\caption{\textbf{NLQ results for Ego4D with MAE trained on Ego4D.}}
\label{tbl:mvp}
\end{table}

\subsection{Additional experiments with pose embeddings and ground-truth pose} \label{sec:supp_pose_experiments}
As noted in \refsec{sec:pose_embed} ground truth pose may be utilized directly by our model, however it is not easily available in real-world egocentric video, {which makes our use of inferred pose embeddings a strength.

\subsubsection{Importance of pose embeddings}
As noted in \refsec{sec:pose_embed} of the main paper, we train pose embeddings to help relate observations by their visual content as well as relative orientation of capture. We measure the effect of pose embeddings on our local state prediction task. Our models see improvements in predicting objects in each of the cardinal directions (26.1 vs. 24.9 mAP) as well as improved object distance prediction (59\% vs 58\%). This improvement is small for predicting objects in the forward view (+0.7 mAP) where scene information is directly visible, but large for other views that need to be inferred: mAP improvements of +2.2 (right), +0.9 (behind) and +1.0 (left).

\subsubsection{Importance of ground truth pose}
In this section, we explore using pose estimates obtained directly from the simulator or using off-the-shelf structure for motion methods. Note that the existing method \SC{EPC} in \reffig{fig:room_pred_results} and \reftbl{tbl:nlq_results} already uses this ground-truth pose information, which other models (including ours) do not have access to.

\textbf{Extracting pose information.}
For HouseTours videos, we run COLMAP~\cite{schoenberger2016sfm}, a structure from motion framework to compute camera-pose information associated with each frame of the video. For this, we first extract frames from each video at 2fps. Then we run COLMAP using a precomputed vocabulary tree file (flickr100K\_words32K). The resulting trajectories are inherently noisy due to the approximate nature of the SfM pipeline and the absence of true camera parameters for such in-the-wild video. We post-process them by removing erroneous pose values (ones that cause jumps in pose atypical to smooth motion).

Overall, COLMAP successfully localizes $\aprox$32 hours of video from 886 houses out of the original 119 hours available in \cite{chang2020semantic}. Some examples of video trajectories are shown in \reffig{fig:supp_video_trajectories}. Comparing these to the simulated trajectories in \reffig{fig:supp_walkthrough_examples}, we see smoother trajectories overall, but with unrealistic jumps in localizations and loop-closure failures. Note that we visualize only the trajectory, not obstacles in the environment, as we do not have access to occupancy maps for the real-world videos.

Note that these videos are tours of indoor spaces with the intention of visually covering a large area. As a result, the camera-wearer moves slowly and smoothly to show parts of the house. Despite this, only a fraction of trajectories can be localized highlighting the difficulty of estimating pose from monocular video. In contrast, the same procedure fails to localize the camera-wearer in Ego4D videos due to rapid head motions and motion blur. On MP3D, pose is available directly from the simulator.

\textbf{Performance with ground-truth pose.}
In \reffig{fig:supp_room_pred_pose_results}, we investigate the role of pose information for \SC{RoomPred} (top) and \SC{NLQ} (bottom), by directly embedding ground-truth pose as part of the input to the baseline and our model. We find that our model can benefit from pose on MP3D, but falls short on HouseTours, due to noise in extracted pose (compared to simulator-provided pose in MP3D). However, our approach (with and without pose) outperforms \SC{EPC}, which explicitly leverages pose both at train and test time.

\begin{figure}[t]
\centering
\includegraphics[width=0.7\columnwidth]{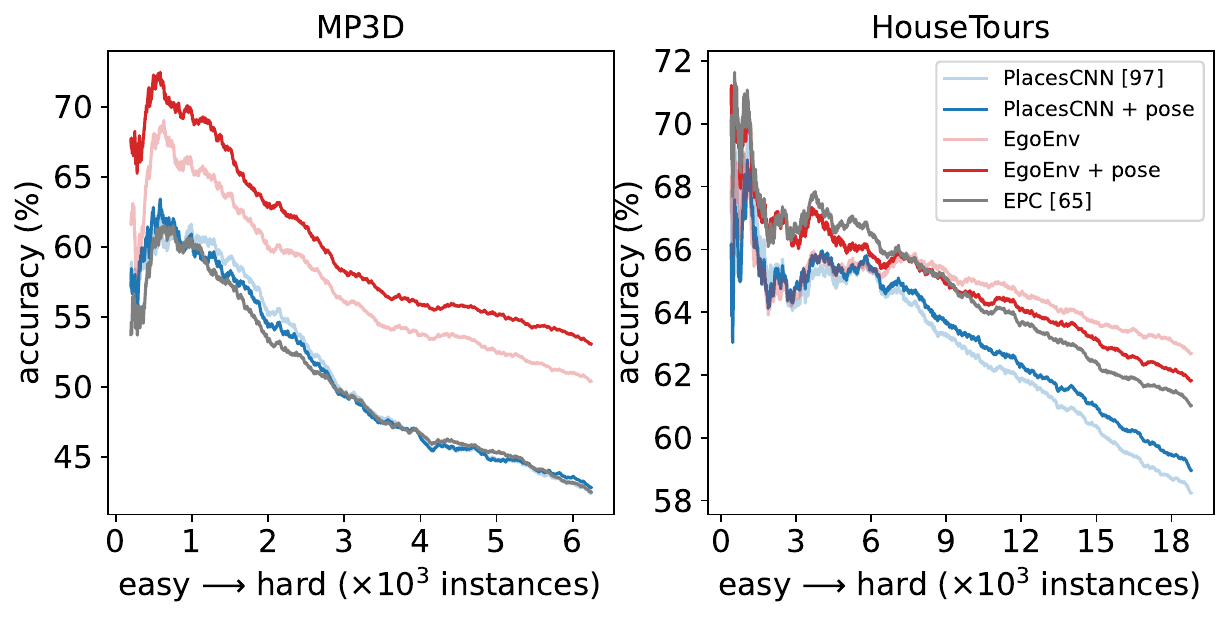}
\small
\centering
\begin{tabular}{|l|ccc|c|ccc|}
\multicolumn{1}{c}{}  &  \multicolumn{3}{c}{MP3D~\cite{chang2017matterport3d}} & \multicolumn{1}{c}{} & \multicolumn{3}{c}{HouseTours~\cite{chang2020semantic}} \\ 
\cline{1-4} \cline{6-8}
\SC{Rank1@m} $\rightarrow$  & @0.3 & @0.5 & \SC{avg} & & @0.3 & @0.5 & \SC{avg} \\ 
\cline{1-4} \cline{6-8}
\SC{VSLNet}~\cite{zhang2020span}
                    & 33.69     & 22.83     & 28.26     &
                    & 42.94     & 27.68     & 35.31     \\
\SC{VSLNet + pose}  & 34.68     & 23.70     & 29.19     &
                    & 43.22     & 25.56     & 34.39     \\
\SC{EPC}~\cite{ramakrishnan2021epc}
                    & 36.85     &  \B{27.64}     & 32.24     &
                    & 43.22     & 27.82    & 35.52     \\
\SC{\Acronym}           & 38.18     & 26.85    & 32.52      &
                    & \B{51.98} & \B{34.18} & \B{43.08} \\
\SC{\Acronym~+ pose}    & \B{38.51} & 27.30 & \B{32.91} &
                    & 46.47 & 30.08 & 38.28 \\
\cline{1-4} \cline{6-8}
\end{tabular}
\caption{\textbf{Task performance using ground-truth pose.} \SC{RoomPred} (top) and \SC{NLQ} (bottom). Our approach (w/ and w/o pose) outperforms baselines. Noisy pose in HouseTours degrades performance. Ego4D videos do not have associated pose.}
\label{fig:supp_room_pred_pose_results}
\end{figure}

\begin{figure*}[t]
\centering
\includegraphics[width=\textwidth]{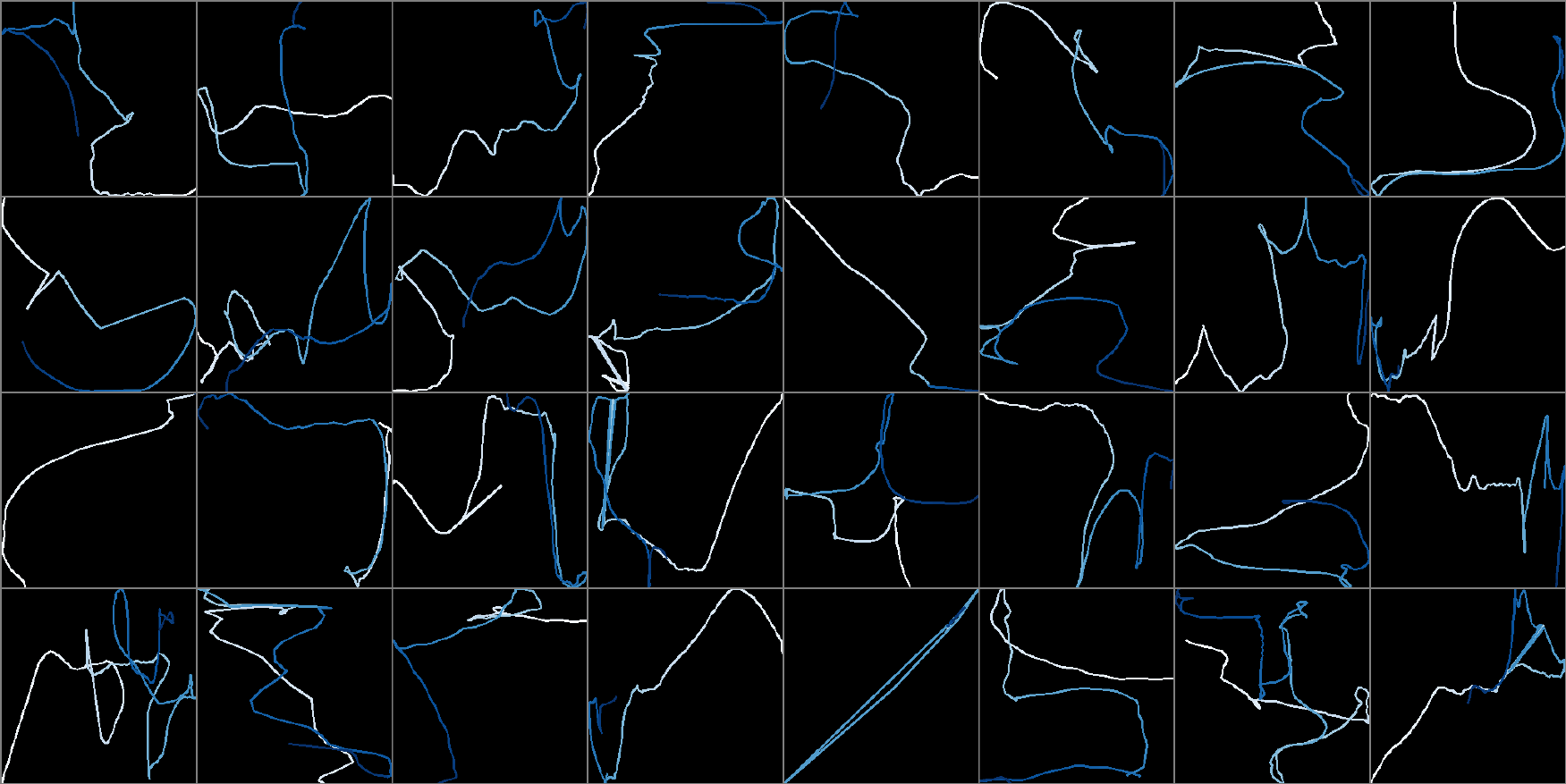}
\vspace*{-0.1in}
\caption{\textbf{Camera-pose for HouseTours from COLMAP.} The blue gradient represents the trajectory from start (white) to end (blue).
}
\label{fig:supp_video_trajectories}
\end{figure*}

\begin{figure*}
\begin{minipage}[c]{\textwidth}
\includegraphics[width=\textwidth]{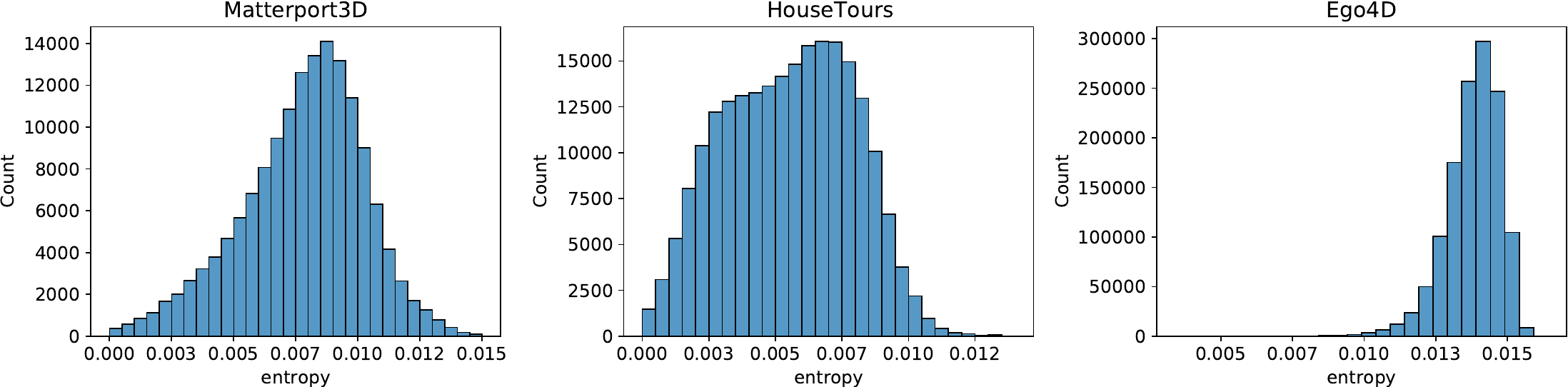}
\end{minipage}
\begin{minipage}[c]{\textwidth}
\includegraphics[width=\textwidth]{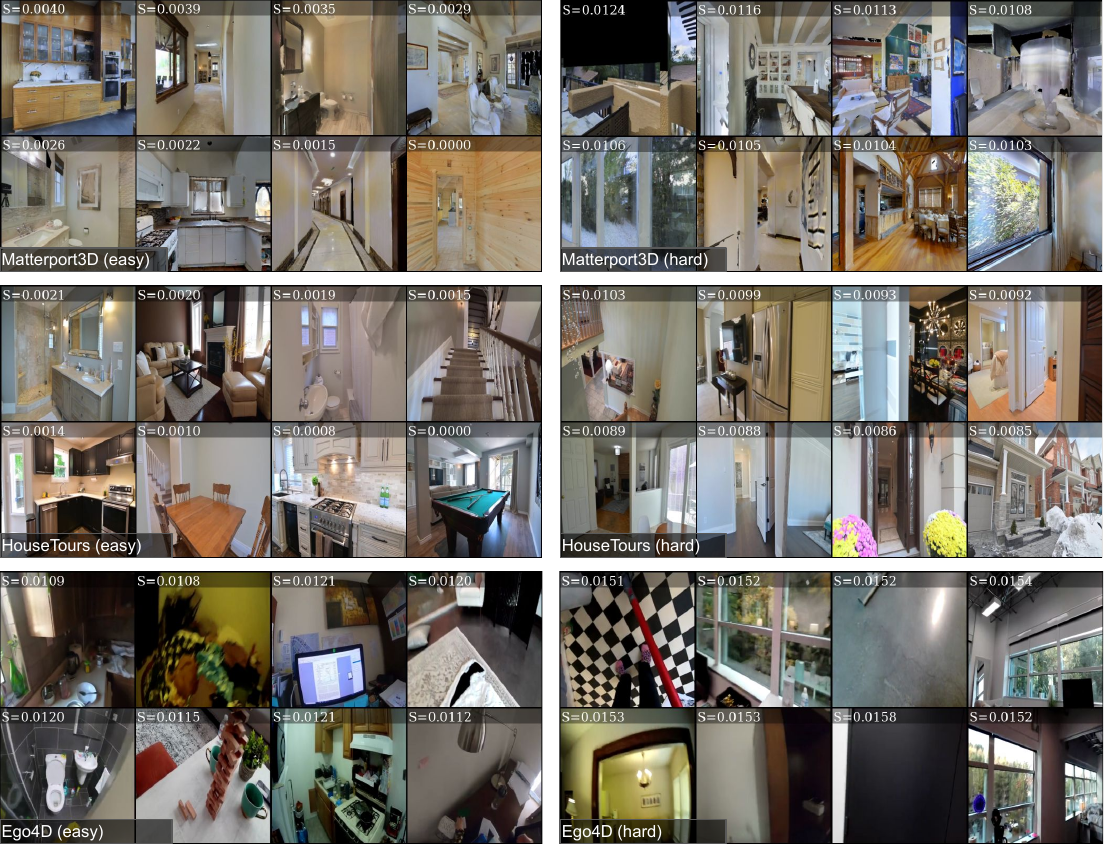}
\caption{\textbf{Illustration of easy vs. hard instances for all datasets.} \textbf{Top panel:} Distribution of entropy scores for \SC{RoomPred} instances. Ego4D instances are skewed towards hard instances due to the egocentric viewpoint and rapid camera and scene motion. \textbf{Bottom panel:} Following \refsec{sec:results_room_pred}, we sort instances as by their entropy score $S$. We show samples from the top and bottom 10\% instances.} 
\label{fig:supp_easy_vs_hard}
\end{minipage}
\end{figure*}

\subsection{EgoEnv integrated into other baseline approaches.} \label{sec:supp_nlq_other_base}
In \reftbl{tbl:nlq_results} in the main paper, we report results on Ego4D using a single architecture and feature combination (VSLNet~\cite{zhang2020span} with SlowFast~\cite{feichtenhofer2019slowfast} features). In \reftbl{tbl:supp_nlq_other_base} we show results with EgoEnv features integrated into other architectures. Our features consistently improve performance across all architectures, highlighting the complementary environment-level information encoded through our approach. 

\begin{table}[t]
\small
\centering
\begin{tabular}{|l|ccc|}
\multicolumn{1}{c}{}  &  \multicolumn{3}{c}{Ego4D~\cite{grauman2022ego4d}} \\
\hline
\SC{Rank1@m} $\rightarrow$ & @0.3 & @0.5 & \SC{avg} \\ \hline
\SC{VSLNet}~\cite{zhang2020span}     
                 & 5.45      & 3.12     & 4.29       \\
\SC{+ \Acronym}  & \B{6.04}  & \B{3.51} & \B{4.77}   \\ \hline
\SC{EgoVLP}~\cite{lin2022egocentric}
                 & \B{10.53} & 5.96     & 8.25       \\
\SC{+ \Acronym}  & 10.51     & \B{6.71} & \B{8.61}   \\ \hline
\SC{ReLER}~\cite{liu2022reler}
                 & 10.79     & \B{6.74} & 8.77       \\
\SC{+ \Acronym}  & \B{11.10} & 6.56     & \B{8.83}   \\ \hline
\SC{ReLER*}      & 13.68     & 8.23     & 10.96      \\
\SC{+ \Acronym}  & \B{14.40} & \B{8.54} & \B{11.47}  \\ \hline
\SC{NAQ~\cite{ramakrishnan2023naq}}
                 & 24.12     & 15.04     & 19.58      \\
\SC{+ \Acronym}  & \B{25.37} & \B{15.33} & \B{20.35}  \\ \hline

\end{tabular}
\caption{\textbf{EgoEnv features with alternate models.} Results on the Ego4D NLQ validation set. \SC{ReLER*} combines EgoVLP~\cite{lin2022egocentric} features with the model from \cite{liu2022reler}.}
\label{tbl:supp_nlq_other_base}
\end{table}

\subsection{Alternate pretraining task formulations.} \label{sec:supp_alternate_tasks}
In \refsec{sec:local_state_pretraining} in the main paper, we introduced our local state prediction task that is used to pretrain our video encoders on simulated walkthrough videos. We investigate alternate pretraining objectives to validate our task choice. We compare against the following:
\begin{itemize}[leftmargin=*]
\itemsep0em 
    \item \textbf{\SC{CardinalObj}} is a variant of our local state tasks where we predict only the object categories in each cardinal direction, but not the distances. 
    \item \textbf{\SC{PoseEmbed}} predicts the relative pose (discretized position and orientation) between every pair of observations in a walkthrough video. This is a sub-component of our full model (\refsec{sec:pose_embed}).
    \item \textbf{\SC{PanoFeat}} directly predicts the image features in each cardinal direction, inspired by prior work on panorama completion~\cite{jayaraman2018learning,song2018im2pano3d,koh2021pathdreamer}.
    \item \textbf{\SC{PanoContrast}} uses noise contrastive estimation (NCE) to train a model to predict image features in each cardinal direction. For positives, we use the true image feature in the corresponding direction direction. For negatives, we use image features from the other 3 cardinal directions, as well as trajectory images from different scenes.
\end{itemize}

Each of these objectives explicitly encodes a combination of semantic and geometric information. For example, \SC{PanoFeat} and \SC{PanoContrast} encodes primarily semantic information as they require reconstruction of image features. \SC{PoseEmbed} encodes primarily geometric information to predict the relative pose between observation pairs. \SC{CardinalObj} encodes semantics from object categories and weak geometric information from their relative orientations.  \reftbl{tbl:supp_alt_pretraining_tasks} highlights the performance of these variants on both downstream tasks. Our approach that requires predicting both object labels, orientations as well as rough distances offers a balance of both cues during pretraining, translating to strong downstream performance.

\begin{table}[t]
\small
\hfill
\subcaptionbox*{}{
\centering
\begin{tabular}{|l|cc|}
\multicolumn{1}{l}{} &  \multicolumn{2}{c}{\SC{RoomPred}} \\ 
\cline{2-3} 
\multicolumn{1}{l|}{} & MP3D    & HT        \\ \hline
\SC{PoseEmbed}      & 38.73     & 59.72     \\
\SC{CardinalObj}    & 49.04     & 61.32     \\
\SC{PanoFeat}       & 48.19     & 62.78     \\
\SC{PanoContrast}   & 47.28     & \B{62.97} \\
\SC{Ours}           & \B{50.40} & 62.68    \\ \hline
\end{tabular}
}
\hfill
\subcaptionbox*{}{
\centering
\begin{tabular}{|l|cc|}
\multicolumn{1}{l}{} & \multicolumn{2}{c}{\SC{NLQ}}    \\ 
\cline{2-3} 
\multicolumn{1}{l|}{} &    MP3D      & HT              \\ \hline
\SC{PoseEmbed}      &    28.46     & 38.56           \\
\SC{CardinalObj}    &    29.32     & 39.34           \\
\SC{PanoFeat}       &    31.34     & 38.06           \\
\SC{PanoContrast}   &    \B{33.22} & 38.56           \\
\SC{Ours}           &    32.51     & \B{43.08}       \\ \hline
\end{tabular}
}
\hfill
\vspace{-0.25in}
\caption{\textbf{Alternate pretraining tasks.} Our local state prediction task offers a good balance of semantic and geometric cues that lead to features with strong downstream performance. For \SC{RoomPred} (left) we report accuracy (\%). For \SC{NLQ} (right) we report mean Rank1@(0.3, 0.5).}
\label{tbl:supp_alt_pretraining_tasks}
\end{table}

\subsection{Memory vs. anticipation during pretraining} \label{sec:supp_mem_vs_ant}
As mentioned in \refsec{sec:pretraining} of the main paper, our pretraining task involves elements of both aggregating information about relevant views spread across the walkthrough, as well as anticipating objects that are rarely (or never seen). We quantify this statement in \reffig{fig:supp_memory_vs_anticipation} where we show the percentage of training instances where objects are rarely seen, for different definitions of rarity. For example, 23\% of training instances involve predicting objects that appear in only $k<4$ frames. 5\% of training instances involve anticipating completely unseen object instances ($k<1$).

\begin{figure}
\centering
\includegraphics[width=0.5\columnwidth]{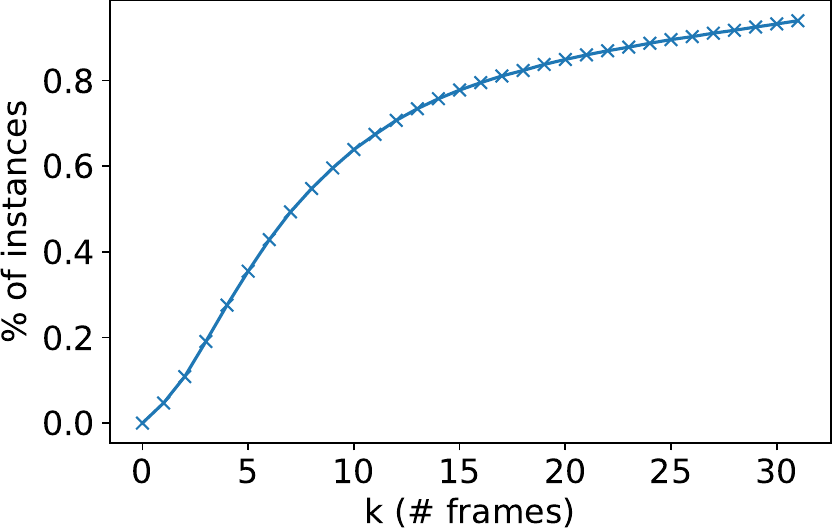}
\caption{\textbf{Percentage of training instances that involve ``rare'' objects.} The x-axis sets a threshold for what is considered rare. For example, 5\% of training instances involve anticipating completely unseen object instances ($k<1$).}
\label{fig:supp_memory_vs_anticipation}
\end{figure}

\subsection{Task-specific pre-training in simulation} \label{sec:supp_task_pretraining}

As mentioned in \refsec{sec:downstream} of the main paper, our goal is to train task-agnostic representations using videos from simulated agents. The end result is a single model that can generate features for multiple tasks (in our experiments, \SC{RoomPred} and \SC{NLQ}}). This is different from traditional sim-to-real approaches where a new dataset needs to be collected for every downstream task, and a separate model has to be trained on it. Such a dataset needs to be well balanced and carefully designed to match the downstream task. Moreover, as tasks are added, new datasets per task need to be created which may be impractical, especially when they require data beyond the simulator’s capability (e.g., simulating human motion, hand-object interaction). 

To investigate further, for the \SC{RoomPred} task, we generate a dataset in simulation that maps trajectory frames to room labels\footnote{We use trajectories from Gibson~\cite{xia2018gibson} scenes as HM3D region annotations are not provided.}. The room categories are estimated directly from the simulator by matching each frame to the nearest navigable point in an annotated room region. We train a ResNet18 model to predict the room category (a six-way classification) and then use features from this model following baselines in \refsec{sec:experiments} (baselines). The new baseline benefits from representations learned for the \emph{same task} --- room prediction --- and on the same volume of simulated training data. 

On MP3D this performs better than the PlacesCNN baseline (43.3 vs. 42.4\%) but is weaker than our model (50.4\%). On HouseTours, it performs worse than PlacesCNN (57.9 vs. 58.2\%) and our approach (62.7\%). The low performance may be attributed to the small label space (only six categories), resulting in features that are not discriminative for the large-scale, diverse data downstream. More generally, the task-specific approach is more susceptible to failures due to the sim-to-real gap, which manifests as a lower performance on real-world video frames from HouseTours.

\section{Ablations experiments and additional visualizations.} \label{sec:supp_ablations}
We present ablation experiments for various model design choices and additional experiments to supplement the discussion in \refsec{sec:ablations} in the main paper.

\subsection{\SC{RoomPred} results with error bars} \label{sec:supp_error_bars}

In \reffig{fig:room_pred_results} of the main paper, we report results over three runs, by aggregating predictions across all three runs, and then sorting them by difficulty (\refsec{sec:results_room_pred}). In \reftbl{tbl:roompredavg}, we show results with standard error \emph{averaged over the three runs} to highlight the variance across approaches. Environment-centric approaches (\SC{EPC}, \SC{TRF}, \SC{EgoEnv}) perform better than other baselines, despite higher variance in accuracy. Our approach is consistently the best amongst these. 

\begin{table}[ht]
\centering
\small
\begin{tabular}{|l|c|c|c|}
\multicolumn{1}{c}{} & \multicolumn{1}{c}{MP3D} & \multicolumn{1}{c}{HouseTours} & \multicolumn{1}{c}{Ego4D} \\
\hline
\textsc{PlacesCNN}     	& 42.39 $\pm$ 0.15 & 58.24 $\pm$ 0.02 & 49.50 $\pm$ 0.20 	\\ 
\textsc{FrameFeat}     	& 42.04 $\pm$ 0.08 & 58.70 $\pm$ 0.10 & 49.34 $\pm$ 0.12 	\\ 
\textsc{ObjFeat}       	& 43.72 $\pm$ 0.06 & 59.02 $\pm$ 0.12 & 48.74 $\pm$ 0.11 	\\ 
\textsc{MAE}           	& 42.79 $\pm$ 0.25 & 58.30 $\pm$ 0.08 & 48.87 $\pm$ 0.08 	\\ 
\textsc{EgoTopo}       	& 41.19 $\pm$ 0.57 & 58.05 $\pm$ 0.15 & 49.42 $\pm$ 0.05 	\\ 
\textsc{EPC}           	& 42.48 $\pm$ 1.12 & 61.02 $\pm$ 0.20 & --          		\\ 
\textsc{Trf (scratch)}	& 43.27 $\pm$ 0.40 & 62.12 $\pm$ 0.14 & 49.65 $\pm$ 0.64 	\\ 
\textsc{EgoEnv}        	& \textbf{50.40 $\pm$ 1.29} & \textbf{62.68 $\pm$ 0.19} & \textbf{51.07 $\pm$ 0.65} 	\\ \hline
\end{tabular}
\caption{\textbf{\textsc{RoomPred} results with error bars across three training runs.}}
\label{tbl:roompredavg}
\end{table}

\subsection{Ablation studies.} \label{sec:supp_extra_ablations}

We perform ablation experiments for several model design choices listed in \refsec{sec:experiments} (experiment setup) of the main paper. All ablation experiments are performed on on validation data splits of MP3D and HouseTours.

\begin{figure*}
\centering
\includegraphics[width=0.8\textwidth]{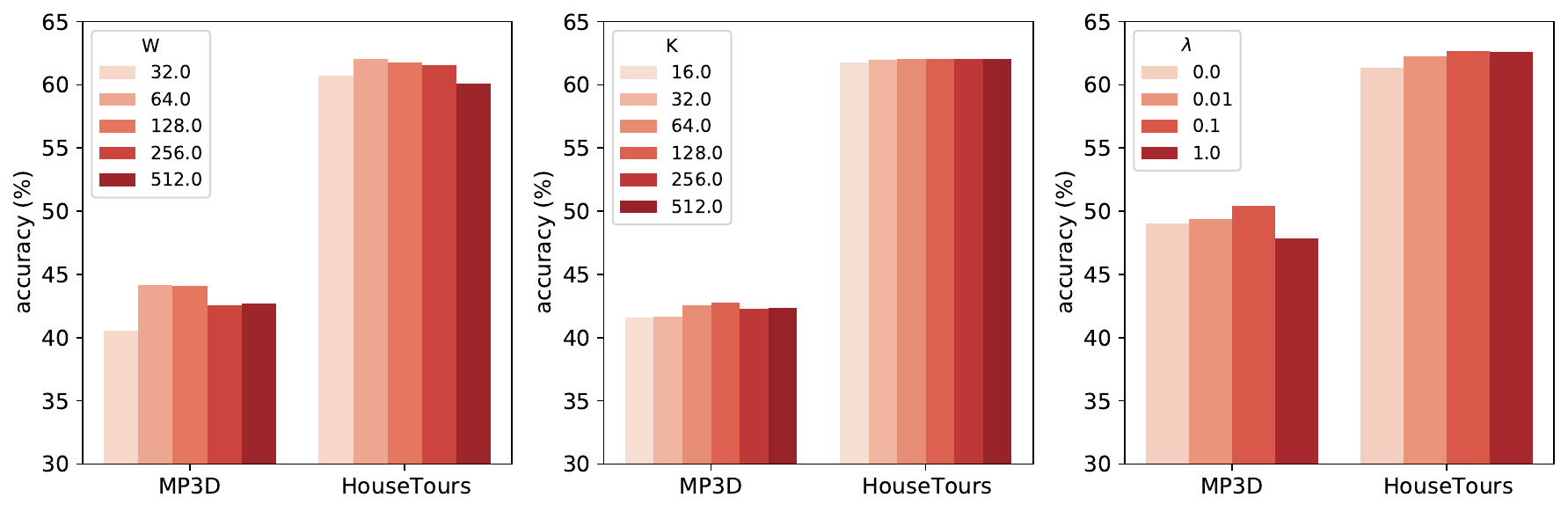}
\includegraphics[width=0.8\textwidth]{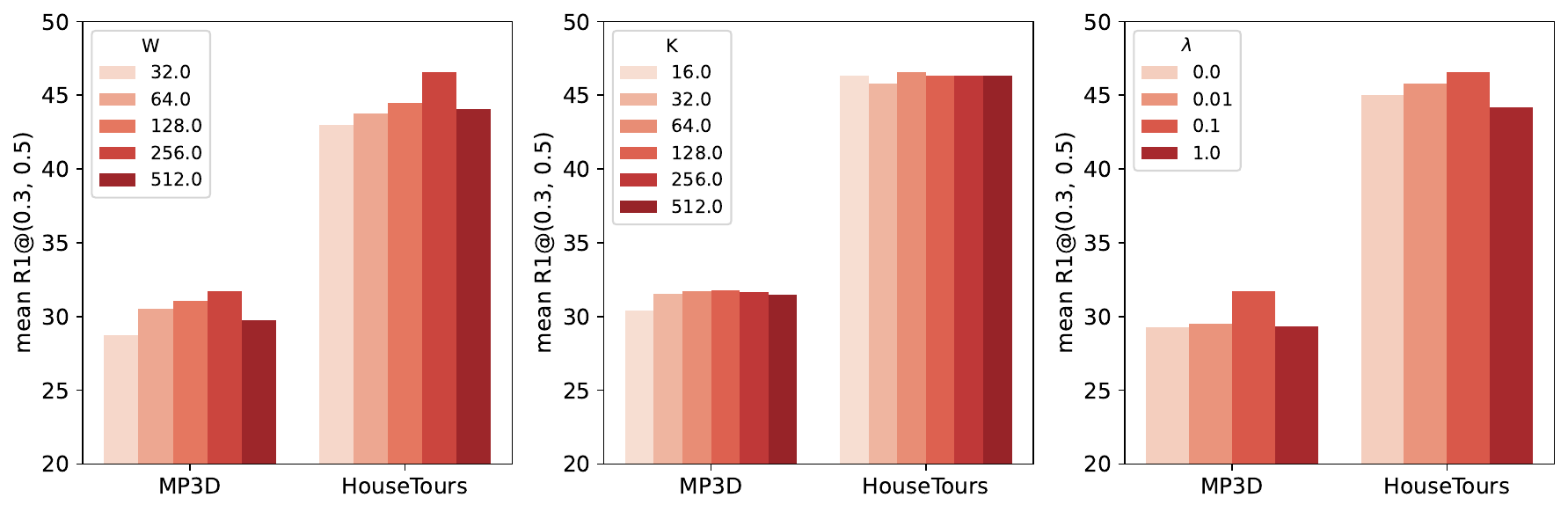}
\caption{\textbf{Ablation experiments on \SC{RoomPred} (top) and \SC{NLQ} (bottom).} We ablate model hyperparameters: window size $W$ (left), memory size $K$ (middle) and loss weight term $\lambda$ (right). See text for analysis.} 
\label{fig:supp_ablations}
\end{figure*}

\textbf{Window size $W$.} The window size controls the density of sampled frames for building our environment memory. Larger windows imply temporally separated frame inputs. Our results are in \reffig{fig:supp_ablations} (left column). We find that $W=64$ is sufficient for localizing the room category for \SC{RoomPred}, while a $W=256$ is best for \SC{NLQ} which requires reasoning over longer horizons.

\textbf{Memory size $K$.} The memory size controls the number of frames sampled from a window of size $W$ for building our environment memory. Our results in \reffig{fig:supp_ablations} (middle column) show the sensitivity of our model to this parameter. On both tasks and datasets, we see only marginal improvements with higher memory sizes, though $K = 64$ results in the best performance.

\textbf{Loss weight term $\lambda$.} $\lambda$ controls the weighting between the object and distance prediction term in the local state prediction loss in \refsec{sec:local_state_pretraining}. Our results in \reffig{fig:supp_ablations} (right column) show that $\lambda = 0.1$ results in the best balance between the two semantic and geometric terms.

\subsection{Additional attention visualizations} \label{sec:supp_attn_viz}
\begin{figure*}[t]
\centering
\includegraphics[width=0.8\textwidth]{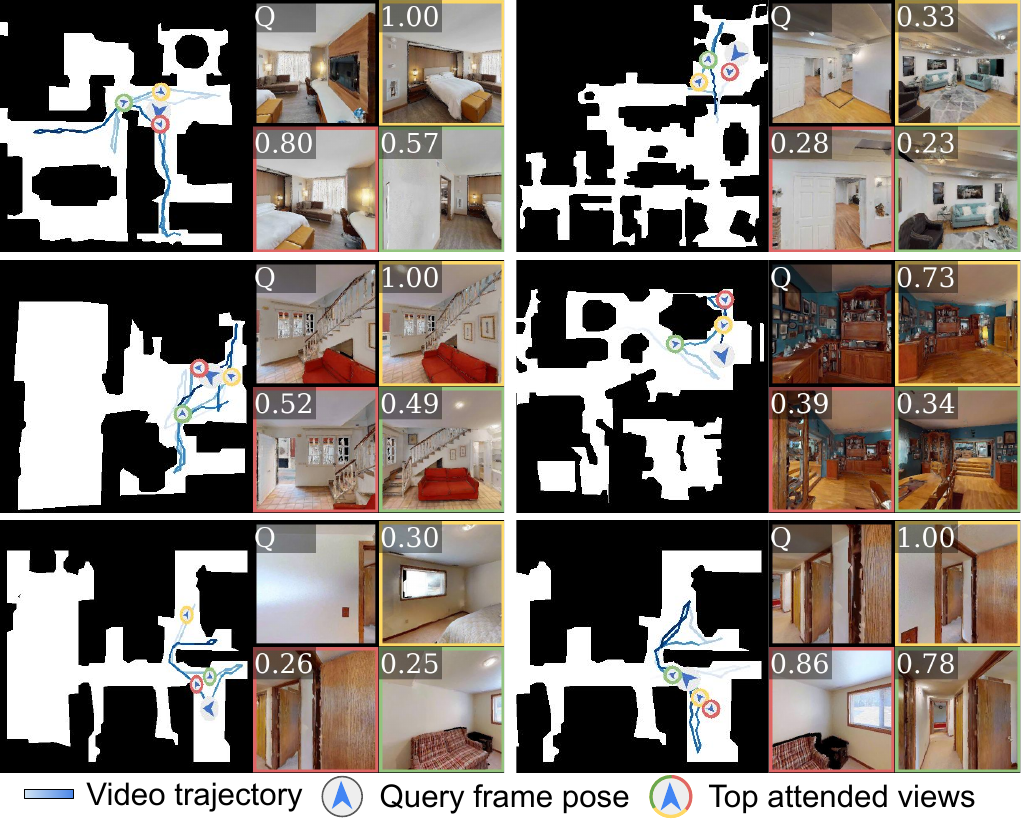}
\vspace*{-0.1in}
\caption{\textbf{Visualized attention weights.} Following \reffig{fig:attn_viz}, the query frame (top left) and top-3 attended views (colored boxes), their positions along the trajectory (colored circles), and their associated attention scores are shown. 
}
\label{fig:supp_attn_viz}
\end{figure*}

We present additional examples visualizing the learned attention values in our transformer decoder model in \reffig{fig:supp_attn_viz} to supplement \reffig{fig:attn_viz} of the main paper. Our model learns to attend to diverse views that are not simply based on temporal adjacency or visual overlap --- they capture the surroundings of the camera-wearer.

\subsection{Easy vs. Hard instances in the \SC{RoomPred} task} \label{sec:supp_easy_vs_hard}
As mentioned in \refsec{sec:results_room_pred} of the main paper, we sort instances for evaluation by difficulty based on the prediction entropy of a pre-trained scene classifier model. In \reffig{fig:supp_easy_vs_hard} (top), we show the distribution of this entropy score across all three datasets. In general, Ego4D contains the hardest instances due to characteristic egocentric motion patterns, while HouseTours contains easier examples where the camera-wearer tends to dwell in one particular location to showcase it. We show examples of easy vs. hard instances in \reffig{fig:supp_easy_vs_hard} (bottom). Note that the figure only shows the center frame of the clip that is used to predict the room label to highlight the difference between easy and hard frames. Our results in \reffig{fig:room_pred_results} highlight the advantage of our approach on these hard instances where environment-level reasoning is essential. See our video in \refsec{sec:supp_video} for more context.

\end{document}